\documentclass[10pt,twocolumn,letterpaper]{article}

\usepackage[pagenumbers]{iccv}
\usepackage{multirow}
\usepackage{multicol}
\usepackage{makecell}
\usepackage{color}
\usepackage{xcolor}
\usepackage{colortbl}
\usepackage{bm}
\usepackage{amsmath}
\usepackage{amssymb}
\usepackage{graphicx}
\usepackage{extarrows}
\usepackage{tcolorbox}
\usepackage{CJK}
\usepackage{hhline}

\definecolor{lightgray}{rgb}{0.88, 0.92, 0.98}
\definecolor{defblue}{rgb}{0.1843, 0.3333, 0.6}
\definecolor{defred}{rgb}{0.88, 0.2510, 0.3294}

\definecolor{green1}{rgb}{ 0.910,  0.953,  0.855}
\definecolor{green2}{rgb}{0.82,  0.902,  0.710}
\definecolor{green3}{rgb}{0.713,  0.903,  0.648}
\definecolor{green4}{rgb}{ 0.725,  0.855,  0.561}

\definecolor{defyellow}{rgb}{1,  0.983,  0.717}
\definecolor{defyellowtext}{rgb}{1,  0.851,  0.438}

\newcommand{\statement}[1]{\noindent\textbf{#1}}
\definecolor{iccvblue}{rgb}{0.21,0.49,0.74}
\usepackage[pagebackref,breaklinks,colorlinks,citecolor=iccvblue]{hyperref}

\def\confYear{2025}

\title{FineCIR: Explicit Parsing of Fine-Grained Modification Semantics for Composed Image Retrieval}

\author{Zixu Li$^{1}$\quad Zhiheng Fu$^{1}$\quad Yupeng Hu$^{1}$\footnote{Corresponding Author.} \quad Zhiwei Chen$^{1}$\quad Haokun Wen$^{2,3}$\quad  Liqiang Nie$^{3}$ \\
$^1$ School of Software, Shandong University \quad \\
$^2$ School of Data Science, City University of Hong Kong\\ 
$^3$ School of Computer Science and Technology, Harbin Institute of Technology (Shenzhen) \quad
\\
{\tt\small \{Lzx, zivchen, fuzhiheng0215\}@mail.sdu.edu.cn} \\ 
{\tt\small huyupeng@sdu.edu.cn,}
{\tt\small \{whenhaokun, nieliqiang\}@gmail.com}
}

\begin{document}
\maketitle
\begin{abstract}
Composed Image Retrieval (CIR) facilitates image retrieval through a multimodal query consisting of a reference image and modification text. The reference image defines the retrieval context, while the modification text specifies desired alterations. However, existing CIR datasets predominantly employ coarse-grained modification text (CoarseMT), which inadequately captures fine-grained retrieval intents. This limitation introduces two key challenges: (1) \textbf{ignoring detailed differences leads to imprecise positive samples}, and (2) \textbf{greater ambiguity arises when retrieving visually similar images}. These issues degrade retrieval accuracy, necessitating manual result filtering or repeated queries. To address these limitations, we develop a robust \textbf{fine-grained CIR data annotation pipeline} that minimizes imprecise positive samples and enhances CIR systems' ability to discern modification intents accurately. Using this pipeline, we refine the FashionIQ and CIRR datasets to create two fine-grained CIR datasets: \textbf{Fine-FashionIQ} and \textbf{Fine-CIRR}. Furthermore, we introduce \textbf{FineCIR}, the first CIR framework explicitly designed to parse the modification text. FineCIR effectively captures fine-grained modification semantics and aligns them with ambiguous visual entities, enhancing retrieval precision. Extensive experiments demonstrate that FineCIR consistently outperforms state-of-the-art CIR baselines on both fine-grained and traditional CIR benchmark datasets. Our FineCIR code and fine-grained CIR datasets are available at \href{https://github.com/SDU-L/FineCIR.git}{https://github.com/SDU-L/FineCIR.git}.
\end{abstract}

\vspace{-1.0em}
\section{Introduction}
\label{sec:intro}

Composed Image Retrieval (CIR) is a crucial paradigm in image retrieval, with widespread applications in fields such as product search~\cite{Product-Search1, Product-Search2} and information retrieval~\cite{imagebased2, textbased2}. Its queries take a multimodal form, comprising a reference image and a modification text, where the modification text articulates the user's modification intent regarding the reference image. The CIR model retrieves the target image that aligns most closely with the modified reference image by analyzing the multimodal query. Due to the unique ability of CIR to convey user modification intent, it has garnered significant attention in recent years~\cite{syncmask, cmap, iudc, alret, manme, dwc}.

According to their respective domains, current mainstream CIR datasets can be categorized into two types: open-domain datasets (CIRR~\cite{cirr}, LaSCo~\cite{case}, SynthTriplets18M~\cite{compodiff-SynthTriplets18M}) and fashion-domain datasets (FashionIQ~\cite{FashionIQ}, Shoes~\cite{shoes}, Fashion200K~\cite{fashion200k}). 
However, these datasets are often annotated using coarse-grained modification text (CoarseMT), which fails to convey users' fine-grained retrieval needs. Due to the following limitations, they may negatively impact training and the CIR process.

 \begin{figure}[t!]
   \begin{center}
   \includegraphics[width=0.98\linewidth]{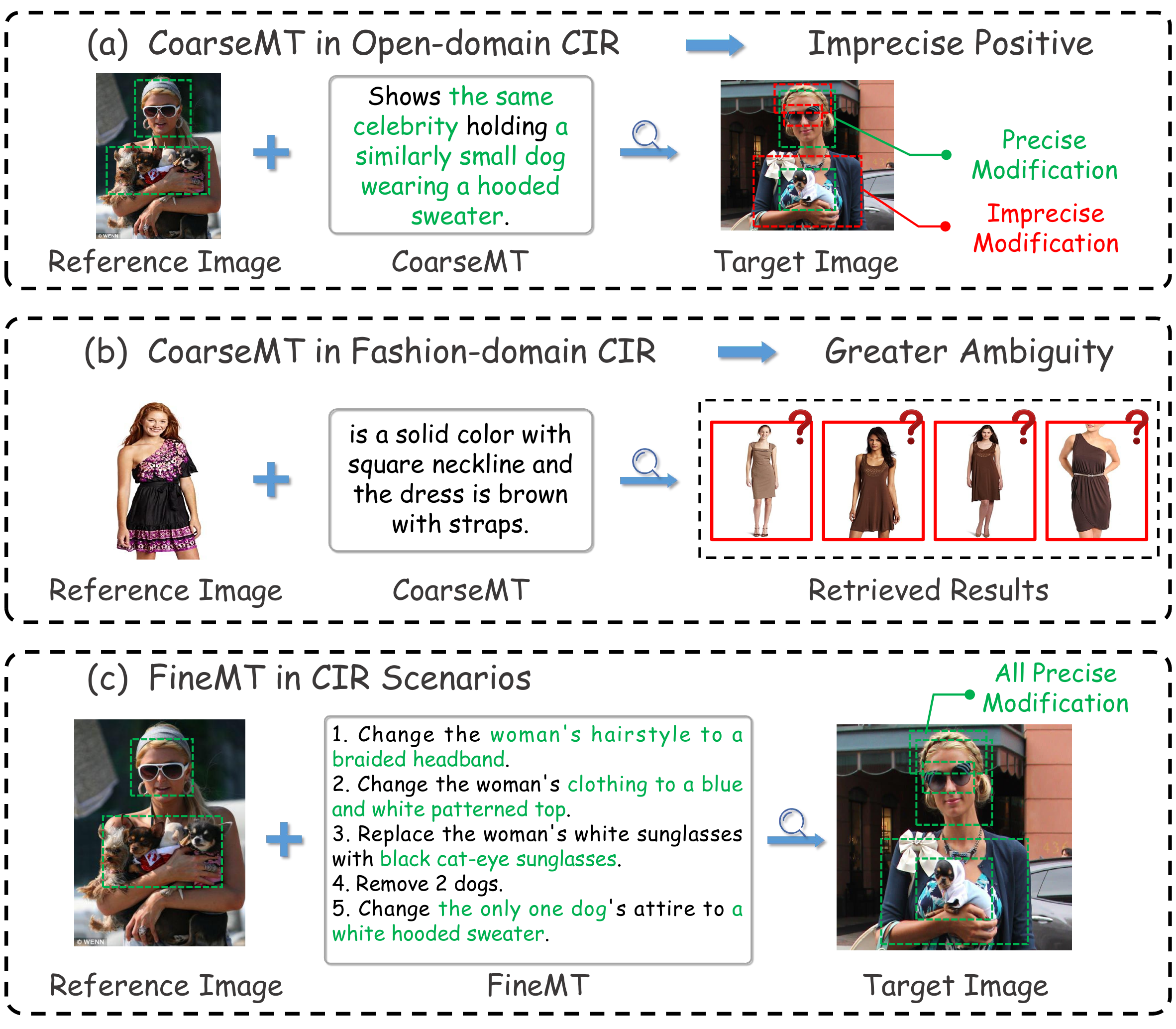}
   \end{center}
   \vspace{-18pt}
   \caption{Problems caused by CoarseMT in (a) Open-domain CIR and (b) Fashion-domain CIR. And (c) illustrates our FineMT in CIR Scenarios.}
   \vspace{-21pt}
   \label{fig:intro-1}
 \end{figure}

\textbf{L1: Ignoring detailed differences leads to imprecise positive samples.} Open-domain datasets are characterized by rich visual content, often exhibiting significant detail differences between the reference image and the target image. As illustrated in~\Cref{fig:intro-1}(a), taking CIRR dataset as an example, the bounding boxes highlight these differences between the reference image and the target image: green boxes indicate the differences mentioned in the CoarseMT, while red boxes indicate the differences that the CoarseMT fails to capture. This suggests that CoarseMT often overlooks fine details, causing deviations from the intended modifications. We define such target images as imprecise positive samples, which are prevalent in existing datasets. When used for training, these samples may bias CIR models toward accepting candidate images with unintended modifications, even if they do not fully align with the multimodal query. This phenomenon reduces retrieval accuracy and forces users into iterative refinement and searches.

\textbf{L2: Greater ambiguity arises when retrieving visually similar images.} Fashion-domain datasets are characterized by limited category diversity and a high prevalence of visually similar images. As shown in~\Cref{fig:intro-1}(b), using FashionIQ-Dress as an example, CoarseMT contains only a dozen words or fewer, lacking detail for precise retrieval. This increases ambiguity in retrieval modeling, making it hard for CIR models to differentiate between similar images. Moreover, real-world applications contain even more visually similar images than small-scale datasets, further complicating the retrieval of user-intended results and increasing the burden of manual filtering or repeated queries.

To address these limitations, we introduce fine-grained CIR. As illustrated in ~\Cref{fig:intro-1}(c), the key distinction between fine-grained CIR and traditional CIR is the replacement of CoarseMT with fine-grained modification text (FineMT), which comprehensively describes the differences between the reference image and the target image. This approach reduces imprecise positive samples and enhances the accuracy of CIR systems in capturing user modification intent (more examples in \textcolor{red}{Appendix}~\ref{sup:A.1}). Specifically, compared to traditional CIR, fine-grained CIR significantly enhances CIR models in multiple aspects, including rate of imprecise positive samples, requirements modeling certainty, and retrieval accuracy, etc (detailed comparison in \textcolor{red}{Appendix}~\ref{sup:A.2}). To advance research in this area, we develop a \textbf{fine-grained CIR data annotation pipeline} comprising three stages: \textit{Data Selection}, \textit{Construction}, and \textit{Quality Check}. This pipeline enables the enrichment of existing CIR datasets with fine-grained annotations. We apply this pipeline to two widely used CIR datasets, FashionIQ and CIRR, resulting in \textbf{Fine-FashionIQ} and \textbf{Fine-CIRR}. These datasets support the development of CIR models and enhance their fine-grained retrieval capabilities.

Existing CIR models primarily rely on implicit methods to interpret modification text, making fine-grained modification recognition challenging. To address this, we propose \textbf{FineCIR}, an explicit parsing network for Fine-grained semantics in Composed Image Retrieval. FineCIR is the first CIR framework that explicitly parses modification semantics, facilitating the structured extraction of entities, attributes, and relations from the modification text. By leveraging explicit parsing, FineCIR enables a precise understanding of FineMT's semantics while aggregating extracted information based on its corresponding entity. We then use these explicitly aggregated entity tokens as guidance to achieve indeterminate relationship alignment between visual entities and FineMT's semantics, thereby completing multimodal feature fusion. Through this explicit parsing mechanism, FineCIR more accurately captures modification intent, ultimately enhancing the performance of fine-grained CIR.

The main contributions of this work are as follows:

\begin{itemize}
	\item We construct a fine-grained CIR data annotation pipeline that minimizes imprecise positive samples, improves the handling of visually similar images, and enhances modification intent recognition in CIR systems.
	\item Using this pipeline, we develop two fine-grained CIR benchmark datasets, Fine-FashionIQ and Fine-CIRR. They've both been rigorously validated through automated and manual processes, ensuring their suitability for fine-grained CIR model development. Furthermore, they are fully open-sourced and publicly available.
	\item We propose FineCIR, the first CIR framework to explicitly parse modification semantics. Extensive experiments on four benchmark datasets demonstrate its superior performance on both traditional CIR and fine-grained CIR.
\end{itemize}

\begin{figure*}[ht]
\vspace{-10pt}
  \begin{center}
  \includegraphics[width=0.92\linewidth]{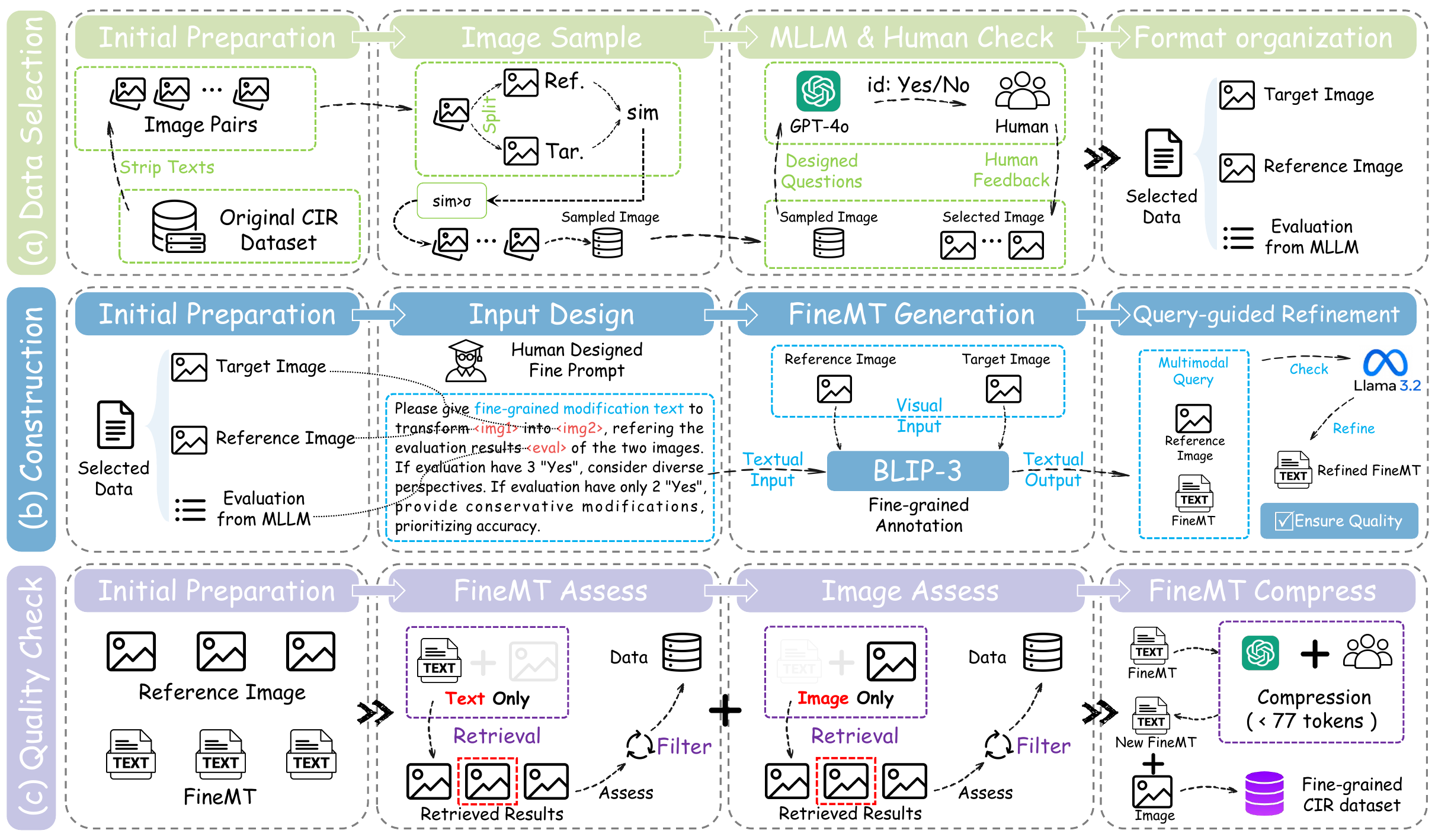}
  \end{center}
  \vspace{-22pt}
  \caption{\small Our fine-grained CIR data annotation pipeline.}
  \vspace{-20pt}
  \label{fig:dataset}
\end{figure*}
\section{Related Work}
\noindent\textbf{Composed Query Datasets}.
Composed query datasets are structured as triplets, each comprising a multimodal query and a target. The query consists of a reference image/video and a modification text. These datasets are widely used in retrieval tasks, including Composed Image Retrieval (CIR)~\cite{tirg, limn, compodiff-SynthTriplets18M, sprc, afce}, Zero-shot CIR (ZS-CIR)~\cite{cirevl, ldre, seize, lincir}, and Composed Video Retrieval (CoVR)~\cite{covr, covr-2, egocvr, covr-cvpr}. However, most existing studies rely on CoarseMT, which lacks precision for fine-grained retrieval. Wu \textit{et al.}~\cite{fdca} proposed generating detailed textual descriptions from video content to advance fine-grained CoVR research. CIR datasets often overlook the limitations of CoarseMT, leading to imprecise positive samples and greater ambiguity when retrieving similar images, hindering CIR model training and retrieval. To address this, we introduce fine-grained CIR, improving model training and enabling CIR systems to accurately interpret FineMT.

\noindent\textbf{Composed Image Retrieval.}
Composed Image Retrieval (CIR) aims to retrieve target images based on a reference image and modification text. According to the utilized backbone, existing CIR methods can be broadly categorized into two streams: traditional models~\cite{tirg, clvcnet} that separately process and compose visual-textual information, and VLP-based models~\cite{limn, sprc, candidate} that embed multimodal queries to unified feature spaces. In addition, due to the high cost of labeling training data for CIR tasks, some researchers have turned to the Zero-shot CIR (ZS-CIR) task. Different from the CIR approach, existing ZS-CIR studies~\cite{lincir, compodiff-SynthTriplets18M} tend to leverage image-caption pairs to train text inversion networks to achieve ZS-CIR. However, due to the high cost of triplet annotation, research on CIR datasets remains relatively underexplored. To address this challenge, we construct a robust fine-grained CIR data annotation pipeline, integrating automated and manual processes for quality assurance. This pipeline facilitates the incorporation of fine-grained annotations into existing CIR datasets, enhancing support for fine-grained CIR.

\section{Fine-grained CIR data annotation pipeline}
\label{sec:benchmark}
As shown in \Cref{fig:dataset}, we constructed a \textbf{fine-grained CIR data annotation pipeline} to enhance fine-grained CIR research. This pipeline comprises three stages. Except for ``Initial Preparation'', each stage goes through 3 steps.
First, the \textit{Data Selection stage} (\Cref{sec:data_collection}) facilitates image pair selection. Second, the \textit{Dataset Construction and Refinement stage} (\Cref{sec:dataset_construction}) generates FineMT and conducts an initial multimodal query quality check. Finally, the \textit{Quality Check stage} (\Cref{sec:quality_check}) verifies triplet data from multiple perspectives to ensure reliability. We applied this pipeline to two benchmark CIR datasets, FashionIQ and CIRR, resulting in two fine-grained CIR datasets (\textbf{Fine-FashionIQ} and \textbf{Fine-CIRR}). They were designed to support CIR model development and enhance fine-grained retrieval performance.

\subsection{Data Selection}
\label{sec:data_collection}
Since existing CIR datasets provide high-quality reference-target image pairs (hereafter referred to as image pairs), we leverage the triplets from the FashionIQ and CIRR datasets as the starting point for dataset construction to minimize overhead. For each triplet, we retain its image pairs and generate fine-grained modification text (FineMT) to describe their differences. As the following fine-grained CIR data annotation pipeline is applicable to various CIR datasets, we do not distinguish between the original dataset names in subsequent sections. To construct a high-quality fine-grained CIR dataset, as shown in~\Cref{fig:dataset}(a), we first filter the raw image pairs.

\statement{\textbf{Image Sample.}}
First, as shown in the ``Image Sample'' step of ~\Cref{fig:dataset}(a), we initially filter image pairs based on semantic similarity. Specifically, we use the pre-trained vision-language model BLIP~\cite{blip} to encode the images and compute the cosine similarity between each pair. A filtering threshold is set to remove image pairs with similarity scores below this threshold. This filtering step accounts for potential mismatches in the original dataset, where some triplets may contain image pairs with excessive scene transitions and deviate from real-world application scenarios, and thus should be excluded. After similarity-based sampling, we obtain the sampled images from the original CIR dataset.

\statement{\textbf{MLLM \& Human Check.}}
As shown the ``MLLM \& Human Check'' step in~\Cref{fig:dataset}(a), after obtaining sampled images, we refine image pairs using MLLM and human verification. Specifically, we design a set of evaluation questions and employ GPT-4o~\cite{gpt4o} to assess each image pair:

\begin{itemize}
	\item Are the contents of the two images related, ensuring a logical and meaningful modification from the reference to the target?
	\item Can the reference image provide valuable information for retrieving the target image?
	\item Are there sufficient modifiable aspects between the two images, and are these modifications feasible and meaningful for fine-grained modification?
\end{itemize}

We instruct GPT-4o (prompts detailed in \textcolor{red}{Appendix}~\ref{sup:B.1}) to provide three ``Yes/No'' responses per image pair, guiding the following actions:
\begin{itemize}
    \item Three ``Yes'' answers $\Rightarrow$ Retained without modification.
    \item Two ``Yes'' answers $\Rightarrow$ Manually reviewed to determine.
    \item One or zero ``Yes'' answers $\Rightarrow$ Discarded.
\end{itemize}

This process ensures that retained image pairs remain relevant and suitable for fine-grained CIR. Through meticulous verification, we obtain the selected images.

\statement{\textbf{Format organization.}}
As shown in the ``Format organization'' step of ~\Cref{fig:dataset}(a), to facilitate subsequent use, we reorganized the data structure, retaining the selected images in the form of triplets: $<$reference image, target image, evaluation from MLLM$>$. This structured format enhances usability for further processing.

\subsection{Dataset Construction and Refinement}
\label{sec:dataset_construction}
As shown in ~\Cref{fig:dataset}(b), based on the dataset obtained from the \textit{Data Selection stage}, we designed a comprehensive dataset construction process. Recognizing the powerful multimodal comprehension capabilities of BLIP-3~\cite{blip3}, we employ it as our primary automatic annotation tool to generate FineMT. Subsequently, we refine the FineMT through a combination of MLLM and human verification.

\statement{\textbf{Input Design.}}
As shown in the ``Input Design'' step of~\Cref{fig:dataset}(b), based on the selected data, we manually designed fine prompt to serve as input for BLIP-3. Specifically, in these prompts, we instruct the model to generate an initial FineMT for each image pair, leveraging both the image pair and the evaluation from MLLM. In this context, $<$img1$>$ denotes the reference image, $<$img2$>$ denotes the target image, and $<$eval$>$ represents the evaluation from MLLM.

\statement{\textbf{FineMT Generation.}}
As shown in the ``FineMT Generation'' step of ~\Cref{fig:dataset}(b), we provide BLIP-3 with the fine prompt (detailed prompt in \textcolor{red}{Appendix}~\ref{sup:B.2}) as textual input and image pairs as visual input. BLIP-3 compares the $<$img1$>$ and $<$img2$>$ while incorporating $<$eval$>$ as a reference to perform the following operations: (1) If the $<$eval$>$ contains three ``Yes'' responses, the model generates FineMT from multiple perspectives. (2) If the $<$eval$>$ contains two ``Yes'' responses, the model conservatively generates several modifications focusing on visual characteristics while ensuring the text remains error-free. Finally, we obtain the textual output (FineMT) from BLIP-3.

\statement{\textbf{Query-guided Refinement.}}
Since BLIP-3's direct output may contain hallucinations, it is necessary to verify the obtained FineMT to eliminate such artifacts. As shown in the ``Query-guided Refinement'' step of~\Cref{fig:dataset}(b), we design specialized prompts (detailed prompt in \textcolor{red}{Appendix}~\ref{sup:B.1}) and utilize Llama3.2~\cite{llama-3} to review all multimodal queries (including the reference image and FineMT). This process will remove the hallucinated content, yielding a refined FineMT.

\subsection{Quality Check}
\label{sec:quality_check}
Through the aforementioned process, we obtain fine-grained multimodal queries. However, we identify two potential issues in the existing data: (1) FineMT contains more detailed information compared to CoarseMT, which may reduce the relevance of the reference image in the CIR process, making the modification text the dominant factor. (2) Most existing CIR datasets construct image pairs based on semantic similarity. In some cases, the reference image alone may be sufficient to accurately retrieve the target image, rendering the modification text less relevant. This contradicts the fundamental objective of CIR, which relies on both textual and visual information. To address these concerns, as shown in ~\Cref{fig:dataset}(c), we utilize feedback from unimodal retrieval results to filter existing triplets from two perspectives: FineMT and reference image.

\statement{\textbf{FineMT Assess.}}
To avoid issue (1), we exclude the reference image and use only FineMT to retrieve the target image. As shown in the ``FineMT Assess'' step of~\Cref{fig:dataset}(c), we employ BLIP~\cite{blip} to extract features from both FineMT and candidate images, then retrieve candidate images using FineMT features. We filter out triplets where the target image appears as the top-1 retrieval result. Next, we conduct a manual assessment, examining the FineMT in relation to the corresponding image pair: (a) If FineMT is overly detailed and includes critical information originally conveyed by the reference image, we refine it using MLLM to ensure that it does not diminish the reference image's contribution. (b) If the image pair exhibits excessive differences and FineMT merely describes the target image rather than its transformation from the reference image, we discard the triplet.

\statement{\textbf{Image Assess.}}
To avoid issue (2), we exclude FineMT and use only the reference image to retrieve the target image. As shown in the ``Image Assess'' step of ~\Cref{fig:dataset}(c), we filter out triplets where the target image appears as the top-1 retrieval result. A manual assessment is then conducted to determine whether to retain these triplets.

\begin{figure*}[ht]
  \vspace{-5pt}
  \begin{center}
  \includegraphics[width=0.85\linewidth]{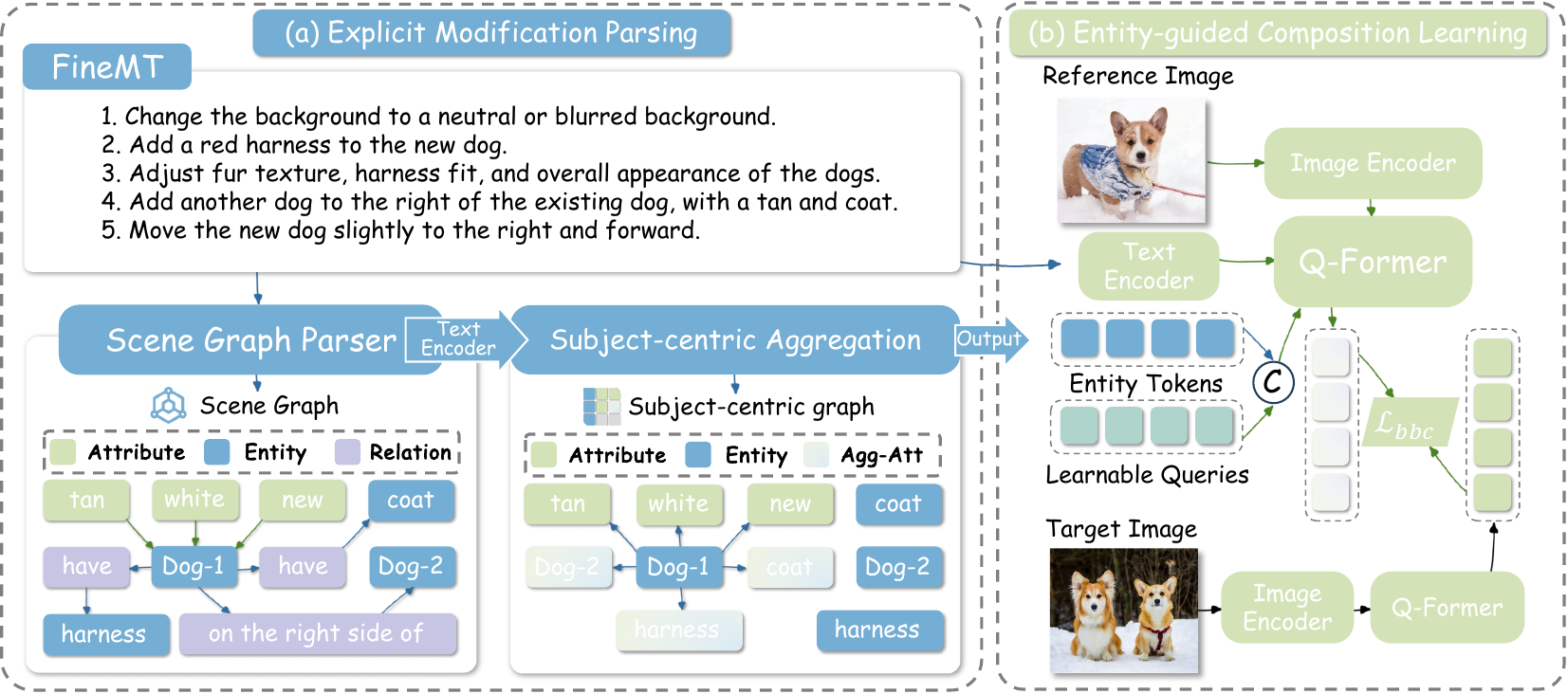}
  \end{center}
  \vspace{-20pt}
  \caption{\small Overall architecture of our proposed FineCIR.}
  \vspace{-20pt}
  \label{fig:framework}
\end{figure*}

\statement{\textbf{FineMT Compress.}}
Finally, we recognize the length constraints imposed by different text encoders, particularly the CLIP encoder, which has a maximum text length of $77$ tokens. To prevent excessively long FineMT from overloading the backbone, we use GPT-4o to verify its length. As shown in the ``FineMT Compress'' step of ~\Cref{fig:dataset}(c), to reduce the workload of manual verification, we first input FineMT exceeding $77$ tokens into GPT-4o (detailed prompt in \textcolor{red}{Appendix}~\ref{sup:B.1}). It will retain key modification information while omitting redundant expressions, producing a compressed version of FineMT. If the length still exceeds $77$ tokens, a manual review is conducted, where human annotators remove irrelevant details while ensuring the completeness of FineMT. Following these steps, we obtain the New FineMT, which, together with the corresponding image pair, forms a triplet. With this, we complete the construction of a fine-grained CIR dataset. In addition, we count the change in query number of Fine-FashionIQ and Fine-CIRR at each step (detail in \textcolor{red}{Appendix}~\ref{sup:B.3}).

\section{Method}
\label{sec:method}
To address the fine-grained CIR, we propose an explicit parsing network of \textbf{Fine}-grained semantics for \textbf{CIR} (\textbf{FineCIR}), which aims to explicitly parse FineMT via the \textit{Scene Graph (SG)} to understand fine-grained modification semantics and achieve indeterminate relationship alignment between visual entities and FineMT's semantics. As illustrated in~\Cref{fig:framework}, FineCIR consists of two main components: (a) \textit{Explicit Modification Parsing (Exparse)} (detailed in~\Cref{sec:Exparse}) and (b) \textit{Entity-guided Composition Learning (Encompose)} (described in~\Cref{sec:Encompose}). In the following, we elaborate on each module of our proposed FineCIR.

\subsection{Preliminaries}
\label{sec:pre}
Given $\mathcal{T}=\left\{\left(x_{r},t_{m}, x_{t}\right)_{n}\right\}_{n=1}^{N}$ as a set of $N$ triplets, where $x_{r}, t_{m}$ and $x_{t}$ refer to the reference image, FineMT and target image, respectively. The designed model aims to learn an embedding space where the multimodal query ($x_{r}, t_{m}$) should be as close as possible to the corresponding target image $x_{t}$, which is formulated as $\mathcal{F}\left(x_r, t_m\right) \rightarrow \mathcal{F}\left(x_t\right)$.

\subsection{Explicit Modification Parsing}
\label{sec:Exparse}
Since FineMT contains a large amount of fine-grained modification semantics, conventional implicit parsing mechanisms struggle to accurately interpret complex fine-grained semantics. To address this issue, we propose the \textit{Explicit Modification Parsing (Exparse)} module. It incorporates a Scene Graph Parser, capable of constructing a scene graph-based structured association framework, which explicitly parses FineMT into an entity-attribute-relation structured semantic representation. Additionally, \textit{Exparse} incorporates Subject-centric Aggregation, which aggregates structured semantics in the scene graph based on subject ownership, serving as guidance for subsequent Entity-guided Composition Learning.

\statement{Scene Graph Parser.} 
To explicitly parse the modification semantics in FineMT, we first construct a scene graph-based structured association framework, extracting a multi-dimensional representation of entity-attribute-relation. Specifically, the scene graph parsing of FineMT \( t_m \) is formulated as follows:

\begin{equation}
    \mathcal{G}(\mathcal{E}, \mathcal{R})=\varPhi_\mathbb{T}(\operatorname{SG\,\,Parser}(t_{m})), 
\label{eq:sg}
\end{equation}

where \( \varPhi_\mathbb{T} \) represents the text encoder of BLIP-2, and \( \mathcal{G} \) denotes the scene graph corresponding to \( t_m \).  \( \mathcal{E} = \{\textbf{E}_1, ..., \textbf{E}_u, ..., \textbf{E}_U\} \) denotes the entity set contained in FineMT, where \( U \) is the entity number. Each entity \( \textbf{E}_u \) has an associated set of attributes, denoted as \( \mathcal{A}_{E_u} = \{\textbf{A}_1, ..., \textbf{A}_k, ..., \textbf{A}_K\} \), where \( K \) is the number of attributes for entity \( \textbf{E}_u \).  
 \( \mathcal{R} = \left\{\left(\textbf{E}_{sub}, \textbf{R}_i, \textbf{E}_{obj}\right)_{i}\right\}_{i=1}^{r} \) denotes the subject-object relation set between entities, where \( r \) is the relation number, and \( \textbf{E}_{sub} \) and \( \textbf{E}_{obj} \) represent the subject and object entities, respectively. \( \textbf{E}, \textbf{A}, \textbf{R} \in \mathbb{R}^{D_\mathbb{T}} \) represent the textual CLS tokens of entities, attributes, and relations, where \( D_{\mathbb{T}} \) is the textual embedding dimension.

\statement{Subject-centric Aggregation.} Given that subject-object relations involve diverse modification semantic operations (\eg, position adjustment and content addition/removal), they pose challenges to accurately modeling modification semantics. To capture the minimal complete semantics of FineMT, we design \textit{Subject-centric Aggregation}, which aggregates structured semantics in the scene graph based on subject ownership, \ie, it consolidates the object relationships associated with the subject, along with its own attributes, into the subject’s semantic representation.

Specifically, based on the subject-object relation set \( \mathcal{R} \) in the scene graph, we reorganize the scene graph \( \mathcal{G} \) into a subject-centric graph \( \mathcal{G}_{sub}=\{\mathcal{E}_{sub}, \mathcal{U}_{sub}\} \), where \( \mathcal{E}_{sub} \) represents the set of all subject entities. \( \mathcal{U}_{sub}\) denotes the union of the subject's attribute set and the associated object set, where the object token integrates the subject-object relation semantics via simple addition, \ie, \textbf{E}+\textbf{R}.

Subsequently, leveraging Graph Attention Networks~\cite{gat, gatv2}, we aggregate the object relations associated with the subject, along with its own attributes, into the subject's semantic representation, formulated as follows:

\begin{equation}
    \textbf{E}_{agg}=\operatorname{GAT}(\mathcal{G}_{sub}), 
\label{eq:gat}
\end{equation}
where \( \textbf{E}_{agg} \in \mathbb{R}^{E \times D_{\mathbb{T}}} \) represents the entity tokens after relational aggregation and \( E \) denotes the subject number.

\subsection{Entity-guided Composition Learning}
\label{sec:Encompose}
Since FineMT contains numerous fine-grained modification details, the same visual entity in the reference image may correspond to multiple modification clauses, and a single modification clause may also correspond to multiple visual entities in the reference image. We refer to this phenomenon as indeterminate relationships. To align these indeterminate relationships and prevent incorrect associations between visual entities and fine-grained modification semantics during multimodal query composition, we propose the \textit{Entity-guided Composition Learning (Encompose)} module. This module leverages explicitly aggregated entity tokens as guidance to identify indeterminate relationships between visual entities in the reference image and fine-grained modification semantics in FineMT. Furthermore, it integrates multimodal query semantics while performing composition learning, promoting the composed multimodal query features close to the target image features. In the following, we detail the module design.

\statement{Feature Extracting.}
Specifically, we first extract the features of the reference image and FineMT. Following previous works~\cite{sprc, covr-2}, we utilize the image encoder of BLIP-2 to extract the visual feature $\textbf{E}_r\!\in\! \mathbb{R}^{C\times D_{\mathbb{I}}}$ of the reference image $x_r$, formulated as,
\begin{equation}
    \textbf{E}_r=\varPhi_\mathbb{I}(x_r), 
\label{eq:summ}
\end{equation}
where $C$ denotes the visual channel and $D_{\mathbb{I}}$ is the visual embedding dimension. $\varPhi_\mathbb{I}$ represent the BLIP-2 image encoder. 
Similarly, we extract the visual feature $\hat{\textbf{E}}_t\!\in\! \mathbb{R}^{C\times D_{\mathbb{I}}}$ of the target image and utilize BLIP-2 text encoder to extract the textual feature $\textbf{E}_m\!\in\! \mathbb{R}^{S\times D_{\mathbb{T}}}$ of the FineMT, where $S$ denotes the textual sequence length.

\noindent \textbf{Entity-guided Composition.}
To explore indeterminate relationships between visual entities in the reference image and fine-grained modification semantics in FineMT, we leverage the entity tokens output from the \textit{Exparse} module as semantic guidance, adaptively aligning the semantic information of the reference image and FineMT.

Specifically, we utilize a set of learnable queries, denoted as $\mathbf{a}_q = \{a_1, ..., a_k\}$. These queries, along with the entity tokens $\textbf{E}_{agg}$ (obtained in~\cref{eq:gat}) and FineMT textual feature $\textbf{E}_m$, are used as the query inputs to the Q-Former.
Since entity tokens explicitly aggregate the subject-centric entity-attribute-relation semantics in FineMT, they enable learnable queries to adaptively aggregate multiple fine-grained modification semantics corresponding to the same visual entity and multiple visual entities corresponding to the same modification semantics within the multimodal query. Moreover, previous studies have shown that Q-Former possesses strong multimodal composition capabilities~\cite{sprc, covr-2}.
Thus, we use the reference image's visual feature $\textbf{E}_r$ as the cross-modal input to Q-Former, facilitating the adaptive alignment of semantic information between the reference image and FineMT, formulated as,
\setlength{\abovedisplayskip}{4pt}
\setlength{\belowdisplayskip}{4pt}
\begin{equation}
        \textbf{E}_c = \operatorname{Q-Former}([\mathbf{a}_q, \textbf{E}_{agg}, \textbf{E}_m], \textbf{E}_r),
        \label{eq:aq}
\end{equation}
where $\textbf{E}_c\!\!\in\!\mathbb{R}^{D}$ denotes the multimodal composed token, \ie, the CLS token of the Q-Former's output, and $D$ denotes the embedding dimension of Q-Former. Similarly, for the target image, we also utilize Q-Former to extract the entity token from its features \( \hat{\textbf{E}}_t \), denoted as \( \textbf{E}_t \in \mathbb{R}^{D} \).

\begin{table*}[ht]
  \centering
  \vspace{-10pt}
        \resizebox{\linewidth}{!}{
        
    \begin{tabular}{l|c|cc|cc|cc|cc|cccc|ccc|c}
        \Xhline{2pt}
    \hline
    \multicolumn{1}{c|}{\multirow{3}{*}{Method}}& \multirow{3}{*}{Year} &  \multicolumn{8}{c|}{Fine-FashionIQ}                              & \multicolumn{8}{c}{Fine-CIRR} \\
\cline{3-18}      &             & \multicolumn{2}{c|}{Dresses} & \multicolumn{2}{c|}{Shirts} & \multicolumn{2}{c|}{Tops\&Tees} & \multicolumn{2}{c|}{Avg} & \multicolumn{4}{c|}{R@k} & \multicolumn{3}{c|}{R$_{subset}$@k} & \multirow{2}{*}{Avg} \\
\cline{3-17}       &            & R@10  & R@50  & R@10  & R@50  & R@10  & R@50  & R@10 & R@50  & k=1   & k=5   & k=10 & k=50 & k=1   & k=2  & k=3 &  \\
    \hline
    \hline
    \rowcolor[rgb]{ .949,  .949,  .949} \multicolumn{18}{c}{\textit{Pre-trained Models}}\\
    Image-Only & \multicolumn{1}{c|}{-} & 10.57      & 25.60      & 16.94      & 31.51      & 14.40      & 30.09      & 13.97      & 29.07      & 8.53      & 30.99      & 44.64      & 76.58      & 21.56      & 41.62      & 61.58 & 26.28 \\
    Text-Only & \multicolumn{1}{c|}{-} & 14.65      & 30.74      & 15.86      & 32.59      & 17.11      & 36.17      & 15.87      & 33.17      & 12.26      & 31.64      & 43.54      & 70.88      & 56.54      & 77.39      & 89.02 & 44.09 \\
    Image+Text & \multicolumn{1}{c|}{-} & 25.38      & 48.11      & 29.55      & 53.11      & 29.14      & 54.02      & 28.02      & 51.75      & 27.00      & 58.17      & 71.72      & 93.91      & 59.77      & 78.91      & 90.28 & 58.97 \\
    \rowcolor[rgb]{ .949,  .949,  .949} \multicolumn{18}{c}{\textit{Existing Methods}}\\
    TIRG~\cite{tirg}  & \multicolumn{1}{c|}{2019} &   7.86  & 16.52  & 10.21  & 16.23  & 9.97  & 22.31  & 9.35  & 18.35  & 8.7l  & 32.41 & 46.59 & 77.35 & 13.68 & 40.23 & 67.88 & 23.05 \\
    CLVC-Net~\cite{clvcnet} & \multicolumn{1}{c|}{2021} &   13.90  & 28.63  & 15.48  & 31.22  & 17.08  & 32.78  & 15.49  & 30.88  & 10.76 & 35.44 & 50.72 & 80.35 & 31.64&	54.94&	74.42	&33.54 \\
    CLIP4CIR~\cite{clip4cir-v2} & \multicolumn{1}{c|}{2022} &    27.58  & 51.12  & 35.62  & 56.91  & 34.77  & 58.43  & 32.66  & 55.49  & 39.48 & 67.36 & 80.69 & 93.87 & 67.45 & 83.10 & 92.34 & 67.41 \\
    TG-CIR~\cite{tgcir} & \multicolumn{1}{c|}{2023} &    33.84  & 59.59  & 48.25  & 69.43  & 46.52  & 70.11  & 42.87  & 66.38  & 46.06 & 77.84 & 87.24 & 97.40 & 72.06 & 87.53 & 94.91 & 74.95 \\
    BLIP4CIR~\cite{blip4cir} & \multicolumn{1}{c|}{2024} &   31.49  & 55.98  & 39.12  & 61.57  & 39.56  & 63.19  & 36.72  & 60.25  & 41.21 & 70.43 & 83.96 & 95.44 & 71.52 & 85.34 & 94.37 & 70.98 \\
    CIReVL~\cite{cirevl} & \multicolumn{1}{c|}{2024} & 24.53  & 50.78  & 31.49  & 51.45  & 36.58  & 57.11  & 30.87  & 53.11  & 33.27 & 62.03 & 73.11 & 89.34 & 66.58 & 83.02 & 93.21 & 64.31 \\
    LinCIR~\cite{lincir} & \multicolumn{1}{c|}{2024} & 37.88  & 61.85  & 43.01  & 63.59  & 47.68  & 70.33  & 42.86  & 65.26  & 31.48 & 62.85 & 72.01 & 88.59 & 66.09 & 81.53 & 92.98 & 64.47 \\
    SEIZE~\cite{seize}& \multicolumn{1}{c|}{2024} & 38.45  & 60.23  & 41.98  & 66.47  & 43.09  & 69.91  & 41.17  & 65.54  & 35.99 & 69.01 & 78.24 & 91.02 & 71.58 & 88.43 & 93.01 & 70.30 \\

    Candidate~\cite{candidate} & \multicolumn{1}{c|}{2024} &    40.69  & 61.56  & 45.63  & 67.40  & 45.09  & 70.37  & 43.80  & 66.44  & 47.96 & 79.29 & 86.01 & 97.28 & 76.89 & 88.03 & 95.62 & 78.09 \\
    SPRC~\cite{sprc}  & \multicolumn{1}{c|}{2024} &     43.89  & 69.18  & \underline{61.86}  & \underline{81.42}  & 60.35  & 80.86  & 55.37  & 77.15  & 53.39  & \underline{81.30}  & 89.31  & 97.58  & \underline{80.65}  & 91.60  & 96.32  & \underline{80.97} \\
    LIMN~\cite{limn}  & \multicolumn{1}{c|}{2024} &    44.04  & 66.31  & 53.61  & 72.90  & 49.97  & 72.62  & 49.20  & 70.61  & 47.35 & 78.44 & 87.31 & 97.30 & 71.69 & 87.23 & 94.20 & 75.07 \\
    CoVR-2~\cite{covr-2} & \multicolumn{1}{c|}{2024} &   \underline{51.21}  & \underline{76.96}  & 60.59  & 78.88  & \underline{60.38}  & \underline{81.27}  & \underline{57.39}  & \underline{79.04}  & \underline{53.48} & 81.12 & \underline{89.87} & \underline{97.66} & 80.34 & \underline{91.73} & \underline{96.55} & 80.73 \\
    \hline
    \hline
            \rowcolor[rgb]{ .851,  .851,  .851}
    \textbf{FineCIR~(Ours)} &    & \textbf{55.29} & \textbf{79.61} & \textbf{64.84} & \textbf{84.82} & \textbf{63.42} & \textbf{83.64} & \textbf{61.18} & \textbf{82.69} & \textbf{57.67} & \textbf{85.69} & \textbf{92.02} & \textbf{98.37} & \textbf{83.77} & \textbf{93.86} & \textbf{97.45} & \textbf{84.73} \\
    \hline
        \Xhline{2pt}
    \end{tabular}%
    }
        \vspace{-10pt}
  \caption{Performance comparison on Fine-FashionIQ and Fine-CIRR relative to R@$K$(\%). The overall best results are in bold, while the best results over baselines are underlined. The Avg metric in Fine-CIRR denotes (R@$5$ + R$_{subset}$@$1$) / 2.}
    \vspace{-18pt}

  \label{tab:main}%
\end{table*}%

\statement{Token Matching Learning}
\label{sec:IEC}
Following previous works~\cite{tgcir, limn, blip4cir}, to promote multimodal composed tokens to align with their corresponding target entity tokens, we apply the universal batch-based classification loss~\cite{val}, formulated as, 

\begin{equation}
\mathcal{L}_{bbc}\!\!=\!\! \frac{1}{B} \sum_{i=1}^{B} -\log \left\{ \frac{\exp \left\{  \operatorname{s} \left( \textbf{E}_{ci} , \textbf{E}_{ti} \right)  / \tau\right\}}{ \sum_{j=1}^{B} \exp \left\{  \operatorname{s} \left( \textbf{E}_{ci} , \textbf{E}_{tj} \right) / \tau \right\}  } \right\},
\label{bbc}
\end{equation}
where as $\textbf{E}_{ci}, \textbf{E}_{ti}$ indicate the composed and target token of the $i$-th triplet, respectively. $\tau$ is the temperature coefficient. $B$ is the batch size.

Finally, the loss function of FineCIR is formulated as,
\begin{equation}
    \mathbf{\Theta^{*}}=
    \underset{\mathbf{\Theta}}{\arg \min } \left( {\mathcal{L}}_{bbc} \right),
    \label{eq:optimization}
    \end{equation}
where $\mathbf{\Theta^{*}}$ is the to-be-optimized parameter for FineCIR.

\section{Experiments}
\label{exp}
In this section, we address the detailed experiments on two fine-grained CIR datasets (Fine-FashionIQ and Fine-CIRR) and two CIR datasets (FashionIQ and CIRR).

\subsection{Experimental Settings}
\textbf{Evaluation.} We choose the recall at rank $K$ (R@$K$) as the evaluation metric, quite similar to the previous CIR task~\cite{val,tgcir,limn}. 
For Fine-FashionIQ and FashionIQ, we employ R@$10$, R@$50$ and their category-wise averages. Similarly, Fine-CIRR and CIRR assessment included R@$k$ ($k\!\!=\!\! 1, 5, 10$), R$_{subset}@k$ ($k\!\!=\!\! 1, 2$) and the average (R@$5$ + R$_{subset}@1$) / $2$. In addition, we provide a description (detailed in \textcolor{red}{Appendix}~\ref{sup:C.1}) and statistical data(detailed in \textcolor{red}{Appendix}~\ref{sup:C.2}) of the above datasets.

\noindent \textbf{Implementation Details.} We utilize BLIP-2~\cite{blip2} as the backbone and train FineCIR using the AdamW optimizer with an initial learning rate of $2e$-$5$. Moreover, we utilize FactualSceneGraph~\cite{factual} as the scene graph parser in the \textit{Explicit Modification Parsing} module. The batch size is set to $128$, and the embedding dimension is set to $256$. 
All experiments are accomplished on a single NVIDIA A$40$ GPU with $48$ GB memory.

\subsection{Method Comparison}
We conduct a comprehensive evaluation of the proposed FineCIR model against several significant baselines on the above four datasets. 
Since the source codes for some methods are not accessible, we select a Vision-Language Pre-training (VLP) model, BLIP~\cite{blip} (for Image-Only, Text-Only, and Image+Text), and some open-source baselines (Candidate~\cite{candidate}, SPRC~\cite{sprc}, LIMN~\cite{limn}, \etc) and the baselines of CIR-related tasks, including ZS-CIR and CoVR. 
We retrain and test all of them according to their original configurations on Fine-FashionIQ and Fine-CIRR. Furthermore, we also evaluate our FineCIR on traditional CIR datasets, \ie, FashionIQ and CIRR, to verify its generalizability. We now detail the analysis as follows.

\statement{Comparison on Fine-grained CIR Datasets.} From the results presented in~\Cref{tab:main}, we can obtain the following conclusions:
\textbf{1)} Our proposed FineCIR achieves superior performance on both Fine-FashionIQ and Fine-CIRR, indicating its excellent ability to comprehend the fine-grained modification semantics.
\textbf{2)} FineCIR demonstrates a significant performance advantage over the baselines, particularly gaining $6.6$\% improvement on Avg.R@$10$ for Fine-FashionIQ. This may be attributed to the enhancements provided by our \textit{Explicit Modification Parsing} module, which improves the model's ability to grasp the complex semantics of FineMT.
\textbf{3)} Previous baselines exhibit suboptimal performance on certain metrics, whereas our FineCIR achieves optimal results on all metrics. This indicates that FineCIR keeps superior generalizability in comprehending fine-grained modifications on fashion and open domains.
\textbf{4)} Text-Only and Image-Only trained with BLIP exhibit significantly lower performance than Image+Text and other VLP-based CIR frameworks, confirming that both the reference image and FineMT in our fine-grained triplets are highly relevant to the multimodal query. 
This prevents a single modality from dominating the compositional process, aligning with the fundamental objective of the CIR task.

\begin{table}[htbp]
  \centering
            \vspace{-8pt}
          \resizebox{0.95\linewidth}{!}{
    \begin{tabular}{ll|cc|c}
    \Xhline{2pt}
    \multicolumn{1}{c|}{\multirow{2}{*}{Method}} & \multicolumn{1}{c|}{\multirow{2}{*}{Year}} & \multicolumn{2}{c|}{FashionIQ} & CIRR \\
\cline{3-5}    \multicolumn{1}{c|}{} &       & \multicolumn{1}{c}{Avg. R@10} & \multicolumn{1}{c|}{Avg. R@50} & Avg \\
    \hline
    \hline
    \multicolumn{1}{l|}{TIRG~\cite{tirg}} & \multicolumn{1}{c|}{2019} & \multicolumn{1}{c}{17.40} & \multicolumn{1}{c|}{37.39} & 35.52  \\
    \multicolumn{1}{l|}{CLVC-Net~\cite{clvcnet}} & \multicolumn{1}{c|}{2021} & \multicolumn{1}{c}{30.70 } & \multicolumn{1}{c|}{58.41 } & - \\
    \multicolumn{1}{l|}{CLIP4CIR~\cite{clip4cir-v2}} & \multicolumn{1}{c|}{2022} & \multicolumn{1}{c}{38.40 } & \multicolumn{1}{c|}{61.74 } & 69.09  \\
    \multicolumn{1}{l|}{TG-CIR~\cite{tgcir}} & \multicolumn{1}{c|}{2023} & \multicolumn{1}{c}{51.32 } & \multicolumn{1}{c|}{73.09 } & 75.57  \\
    \multicolumn{1}{l|}{BLIP4CIR~\cite{blip4cir}} & \multicolumn{1}{c|}{2024} & \multicolumn{1}{c}{42.63 } & \multicolumn{1}{c|}{66.79 } & 72.08  \\
    \multicolumn{1}{l|}{CoVR-2~\cite{covr-2}} & \multicolumn{1}{c|}{2024} & \multicolumn{1}{c}{49.96 } & \multicolumn{1}{c|}{71.17 } & 78.92  \\
    \multicolumn{1}{l|}{Candidate~\cite{candidate}} & \multicolumn{1}{c|}{2024} & \multicolumn{1}{c}{51.17 } & \multicolumn{1}{c|}{73.13 } & 80.90  \\
    \multicolumn{1}{l|}{SPRC~\cite{sprc}} & \multicolumn{1}{c|}{2024} & \multicolumn{1}{c}{54.72 } & \multicolumn{1}{c|}{74.97 } & \underline{81.39}  \\
    \multicolumn{1}{l|}{LIMN~\cite{limn}} & \multicolumn{1}{c|}{2024} & \multicolumn{1}{c}{\underline{55.91} } & \multicolumn{1}{c|}{\underline{77.82} } & 72.19  \\
    \hline
    \hline
                \rowcolor[rgb]{ .851,  .851,  .851}
    \multicolumn{2}{l|}{\textbf{FineCIR}} &  \textbf{59.86}	& \textbf{80.43}   & \textbf{81.63}  \\
    \Xhline{2pt}
    \end{tabular}%
    }
          \vspace{-8pt}
      \caption{Performance comparison on original FashionIQ and CIRR relative to R@$K$(\%). The overall best results are in bold, while the best results over baselines are underlined. The Avg metric in CIRR denotes (R@$5$ + R$_{subset}$@$1$) / 2.}
      \vspace{-12pt}
  \label{tab:cir}%
\end{table}%

\statement{Comparison on Traditional CIR Datasets.}
Due to space constraints, we present the averaged results on the FashionIQ and CIRR datasets in~\Cref{tab:cir}, while the full results are provided in \textcolor{red}{Appendix}~\ref{sup:D}. 
The results indicate that our proposed FineCIR still achieves the best performance on traditional CIR datasets, demonstrating its ability to effectively comprehend coarse-grained modification semantics. This confirms its strong generalization capability across both coarse and fine granularities.

\subsection{Ablation Study}
\label{sec:abla}
In this section, we introduce the ablation study of our proposed FineCIR with different variants, as shown in~\Cref{tab:abla}. The compared variants are as follows.

\begin{itemize}
    \item \textbf{w/o SG.} We replace the scene graph with the original FineMT for subject interaction to verify the scene graph's ability to comprehend fine-grained modifications.
    \item \textbf{w/o SC-Agg.} We remove the \textit{Subject-centric Aggregation} module and use mean pooling on entity features to verify the effectiveness of learning subject-object relations.
    \item \textbf{w/ SG\_AMRBART} \& \textbf{w/ SG\_Stanford.} To verify the robustness of FineCIR, we replace the FactualSceneGraph with AMRBART~\cite{AMRBART} and StanfordSceneGraph~\cite{stanfordSG}.
    \item \textbf{w/o Entity\_Guide.} To verify the efficacy of entity tokens in learning indeterminate relations, we replace them with the mean pooling feature of all tokens in the scene graph.
    \item \textbf{w/o Q-Former.} To verify the role of Q-Former, we replace it with an MLP for composition.
    \item \textbf{w/o BBC.} To verify the point of BBC loss, we replace it with the commonly used KL divergence loss for this task~\cite{tgcir,clvcnet}, which is also used to optimize the distribution of multi-modal queries and target images.
\end{itemize}

\begin{table}[t!]
  \centering

          \resizebox{0.94\linewidth}{!}{
    \begin{tabular}{l|cc|cc|cc}
    \Xhline{2pt}
    \multicolumn{1}{c|}{\multirow{2}{*}{Method}} & \multicolumn{4}{c|}{Fine-FashionIQ} & \multicolumn{2}{c}{Fine-CIRR} \\
\cline{2-7}    \multicolumn{1}{c|}{} & R@$10$ & \multicolumn{1}{c|}{$\Delta$} & R@$50$ & \multicolumn{1}{c|}{$\Delta$} & Avg   & $\Delta$ \\
    \hline
    \hline
    \rowcolor[rgb]{ .949,  .949,  .949} \multicolumn{7}{c}{\textit{Explicit Modification Parsing (Exparse)}}\\
    w/o SG & 57.33  & \cellcolor{green4}-3.85  & 77.77  & \cellcolor{green4}-4.92  & 81.12  & \cellcolor{green4}-3.61  \\
    w/o SC-Agg & 59.01  & \cellcolor{green3}-2.17  & 79.29  & \cellcolor{green3}-3.40  & 82.81  & \cellcolor{green3}-1.92  \\
    w/ SG\_AMRBART & 60.76  & \cellcolor{green1}-0.42  & 82.25  & \cellcolor{green1}-0.44  & 84.24  & \cellcolor{green1}-0.49  \\
    w/ SG\_Stanford & 60.11  & \cellcolor{green2}-1.07  & 81.96  & \cellcolor{green2}-0.73  & 84.01  & \cellcolor{green2}-0.72  \\
    \rowcolor[rgb]{ .949,  .949,  .949} \multicolumn{7}{c}{\textit{Entity-guided Composition Learning (Encompose)}}\\
    w/o Entity\_Guide & 59.87  & \cellcolor{green2}-1.31  & 79.87  & \cellcolor{green2}-2.82  & 83.56  & \cellcolor{green2}-1.17  \\
    w/o Q-Former & 59.63   & \cellcolor{green4}-1.55   & 79.98   & \cellcolor{green4}-2.71   & 82.65   & \cellcolor{green4}-2.08   \\
    \rowcolor[rgb]{ .949,  .949,  .949} \multicolumn{7}{c}{\textit{Loss Function}}\\

    w/o BBC &    58.45    &  \cellcolor{green2} -2.73    &    79.80    & \cellcolor{green2} -2.89      &   81.14 	 & \cellcolor{green2} -3.59    \\
    \hline
    \hline
    \multicolumn{1}{l|}{\textbf{FineCIR~(Ours)}} 
    &  \textbf{61.18}
    & \multicolumn{1}{c|}{\cellcolor[rgb]{ 1,  0.983,  0.717}   -0.00} 
    &  \textbf{82.69 }     
    & \multicolumn{1}{c|}{\cellcolor[rgb]{ 1,  0.983,  0.717}   -0.00}          
    &  \textbf{84.73 }      
    & \cellcolor[rgb]{ 1,  0.983,  0.717}   -0.00         \\
    \Xhline{2pt}
    \end{tabular}%
    }
    \vspace{-5pt}
      \caption{Ablation study on Fine-FashionIQ and Fine-CIRR datasets. We compute Avg-R@10, R@50 for Fine-FashionIQ, and Avg (mean of R@$5$ and R$_{subset}$@$1$) for Fine-CIRR, respectively.}
    \vspace{-14pt}
  \label{tab:abla}%
\end{table}%

From the ablation results of FineCIR in ~\Cref{tab:abla}, we can obtain the following observations.
\textbf{1)} The performance drop of \textbf{w/o SG} is most significant, indicating the superiority of the scene graph in comprehending fine-grained modification semantics. With the aid of scene graph, FineCIR can more precisely comprehend the modification details in FineMT.
\textbf{2)} \textbf{w/o SC-Agg} outperforms \textbf{w/o SG} but still shows a noticeable performance drop, indicating that subject-centric aggregation benefits in learning fine-grained subject modification relationships.
\textbf{3)} \textbf{w/ SG\_AMRBART} and \textbf{w/ SG\_Stanford} both outperform \textbf{w/o SC-Agg}, with only a slight performance drop. This indicates the necessity of scene graphs for fine-grained modification learning and demonstrates that FineCIR's subsequent architecture is robust in learning relationships from various scene graph representations.
\textbf{4)} \textbf{w/o Entity\_Guide} underperforms compared to FineCIR, implying that entity tokens play a crucial role in guiding the multimodal composition process to effectively learn indeterminate relationships.
\textbf{5)} \textbf{w/o Q-Former} performs worse than FineCIR but better than \textbf{w/o SG}, indicating that Q-Former possesses superior multimodal composition capabilities, while also highlighting the indispensability of scene graphs in comprehending fine-grained modifications.
\textbf{6)} \textbf{w/o BBC} underperforms compared to FineCIR, implying that BBC loss is more effective in optimizing metric learning.

\subsection{Case Study}
\label{exp:case_study}
\begin{figure}[t]
     \vspace{-4pt}
  \centering
  \includegraphics[width=\linewidth]{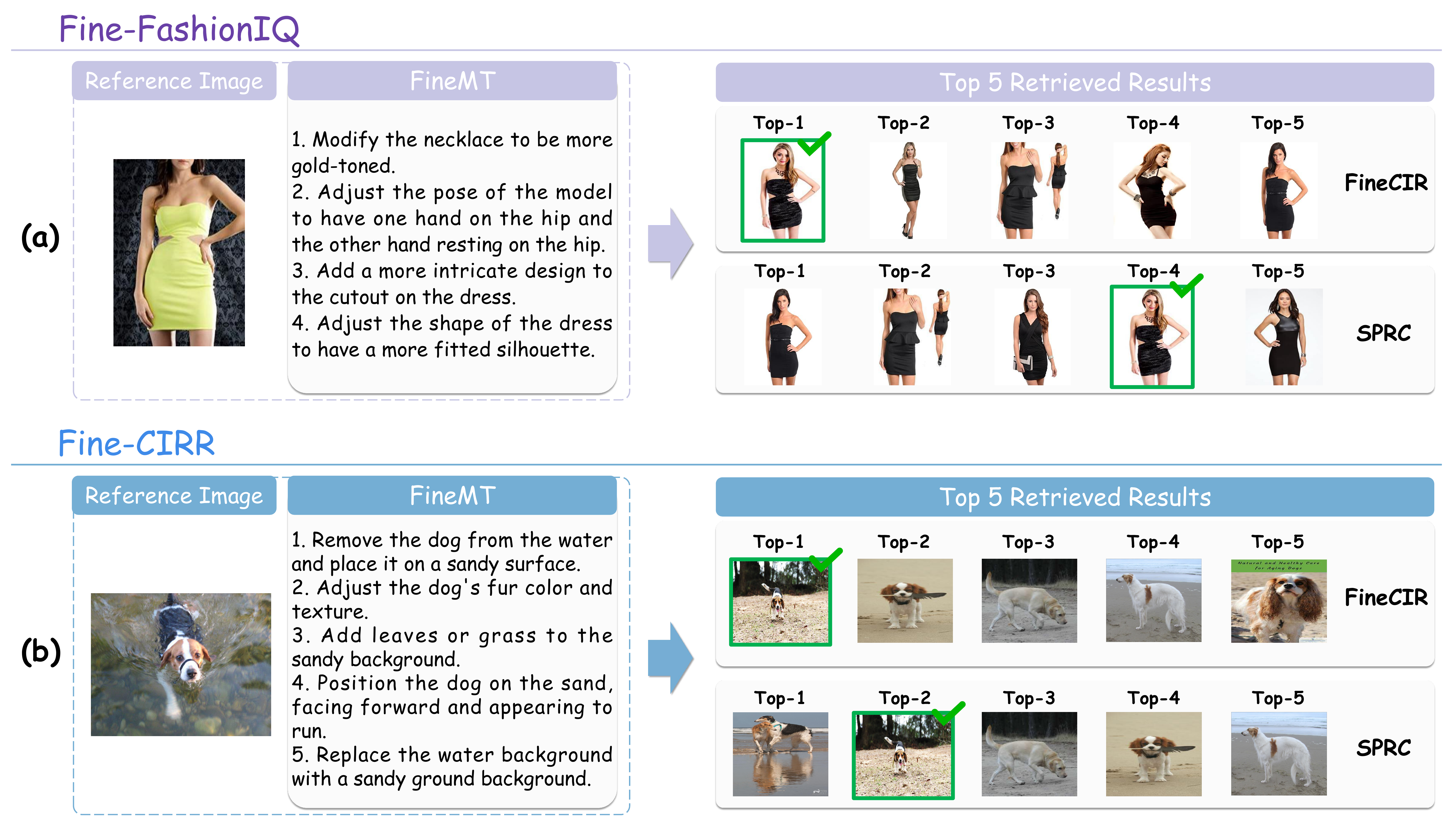}
  \vspace{-18pt}
  \caption{\small Qualitative examples of our proposed FineCIR compared to the sub-optimal CIR model SPRC.}
    \vspace{-18pt}
  \label{fig:case_exp}
\end{figure}

To qualitatively evaluate the performance of our proposed FineCIR in fine-grained CIR scenarios, we present comparative examples against the sub-optimal CIR model SPRC~\cite{sprc} in Figure~\ref{fig:case_exp}. On the Fine-FashionIQ dataset, as shown in Figure~\ref{fig:case_exp}(a), FineCIR retrieves the target image at Top-1, while SPRC ranks it at Top-4, highlighting FineCIR's superior understanding of fine-grained modification texts (FineMT). 
Similarly, on the Fine-CIRR dataset, as shown in Figure~\ref{fig:case_exp}(b), FineCIR captures modifications like repositioning subjects and background changes, ranking the target image at Top-1, compared to SPRC's Top-2, indicating that FineCIR has a stronger capacity to prioritize fine-grained scenarios. 
These qualitative results demonstrate FineCIR's effectiveness in modeling fine-grained semantics across both fashion and open-domain datasets.

\subsection{More Experiments}
Due to space constraints, \textcolor{red}{Appendix}~\ref{sup:E} compares the training and inference efficiency of FineCIR and SOTA, further demonstrating the practicality of FineCIR. Furthermore, \textcolor{red}{Appendix}~\ref{sup:F} provides additional case studies.
\section{Conclusion}
\vspace{-3pt}
In this work, we developed a fine-grained CIR data annotation pipeline and constructed two corresponding datasets that reduced imprecise positive samples and enhanced CIR systems' ability to accurately capture user modification intent. Additionally, we proposed FineCIR, the first CIR framework to explicitly parse modification semantics. Extensive experiments exhibited that FineCIR outperformed existing models in both fine-grained and traditional CIR, highlighting its superiority and generalization capability.

{
    \small
    \bibliographystyle{ieeenat_fullname}
    \bibliography{main}
}

\clearpage
\appendix
\setcounter{page}{1}

\maketitlesupplementary
\noindent This is the supplementary material of ``FineCIR: Explicit Parsing of Fine-Grained Modification Semantics for Composed Image Retrieval''. 
Specifically, Section~\ref{sup:A} illustrates detailed information about the difference between fine-grained CIR and the traditional CIR. 
Section~\ref{sup:B} shows the details of dataset construction, including various prompts for the LLM we utilized in several steps.
Section~\ref{sup:C} presents statistics of the proposed two datasets and compares them with the original CIR datasets.
Section~\ref{sup:D} shows a comprehensive evaluation of the proposed FineCIR model against several significant baselines on the traditional CIR datasets FashionIQ and CIRR.
In Section~\ref{sup:E} we determine the computational costs of our proposed FineCIR, and make comparison with the sub-optimal model.
Section~\ref{sup:F} presents a case study on our proposed fine-graind CIR datasets, Fine-FashionIQ and Fine-CIRR.

\begin{figure*}[htbp]
  \centering
  \includegraphics[width=\linewidth]{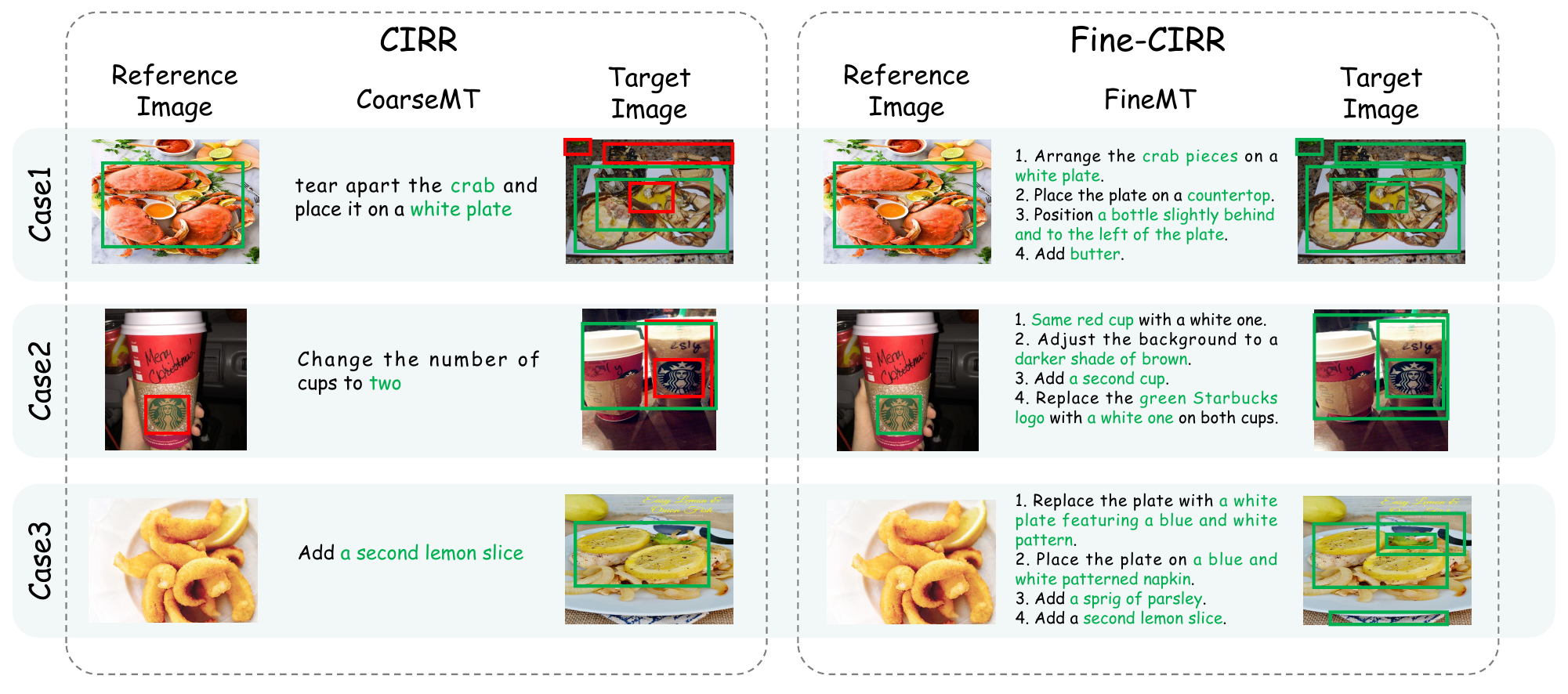}
  \caption{\small Capability of reducing the imprecise positive samples by using the fine-grained modification text (FineMT).}
  \label{fig:false-positive}
\end{figure*}
\section{Fine-grained CIR VS. Traditional CIR}
\label{sup:A}

\subsection{FineMT in Solving CoarseMT's Problems}
\label{sup:A.1}

\statement{Reducing the Imprecise Positive Samples.}
Ground-truth labeling in CIR datasets is frequently inadequate due to the presence of numerous visually similar images and the limited expressive capability of modification text. 
For a given multimodal query, there may exist candidate images that, despite subtle differences from the labeled ground truth, actually align better with the query requirements compared to the target image.
In such cases, these candidate images should ideally be regarded as ground-truth. However, target images that align less effectively with query requirements are often still labeled as positive samples, leading to what we term false-positive samples.

To demonstrate the effectiveness of our proposed fine-grained datasets in reducing the imprecise positive samples, we conducted a comparative experiment between the original coarse modification text (CoarseMT) and our fine-grained modification text (FineMT), as depicted in Figure~\ref{fig:false-positive}. 
In the figure, correctly attended modifications are indicated with green boxes, while incorrectly attended modifications are marked in red.
For example, in Case 1, FineMT covers a more comprehensive set of subjects from multiple perspectives, explicitly including the ``crab'', ``plate'', ``bottle'', and even the ``butter''. Conversely, CoarseMT narrowly focuses only on the ``crab'' and ``plate'', neglecting several other key objects. In situations like these, when a candidate image meets the query criteria more precisely and contains more distinguishing features than the designated target image, the target image can inadvertently be an imprecise positive sample.
By specifying more detailed and comprehensive modification requirements, FineMT significantly reduces the likelihood that suitable candidate images are incorrectly labeled as positive samples. Consequently, this ensures that the target image is accurately classified as a true-positive, effectively mitigating the risk of false-positive classifications.

\statement{Recognizing Visually Similar Images.}
Ground-truth labeling in CIR datasets is often insufficient due to numerous visually similar images and the limited descriptive capability of modification text. For a given multimodal query, candidate images that subtly differ from the ground-truth yet fulfill the query are often visually similar to the target image, complicating the retrieval process by making it challenging to distinguish between visually similar images.

\begin{figure*}[ht]
  \centering
  \includegraphics[width=\linewidth]{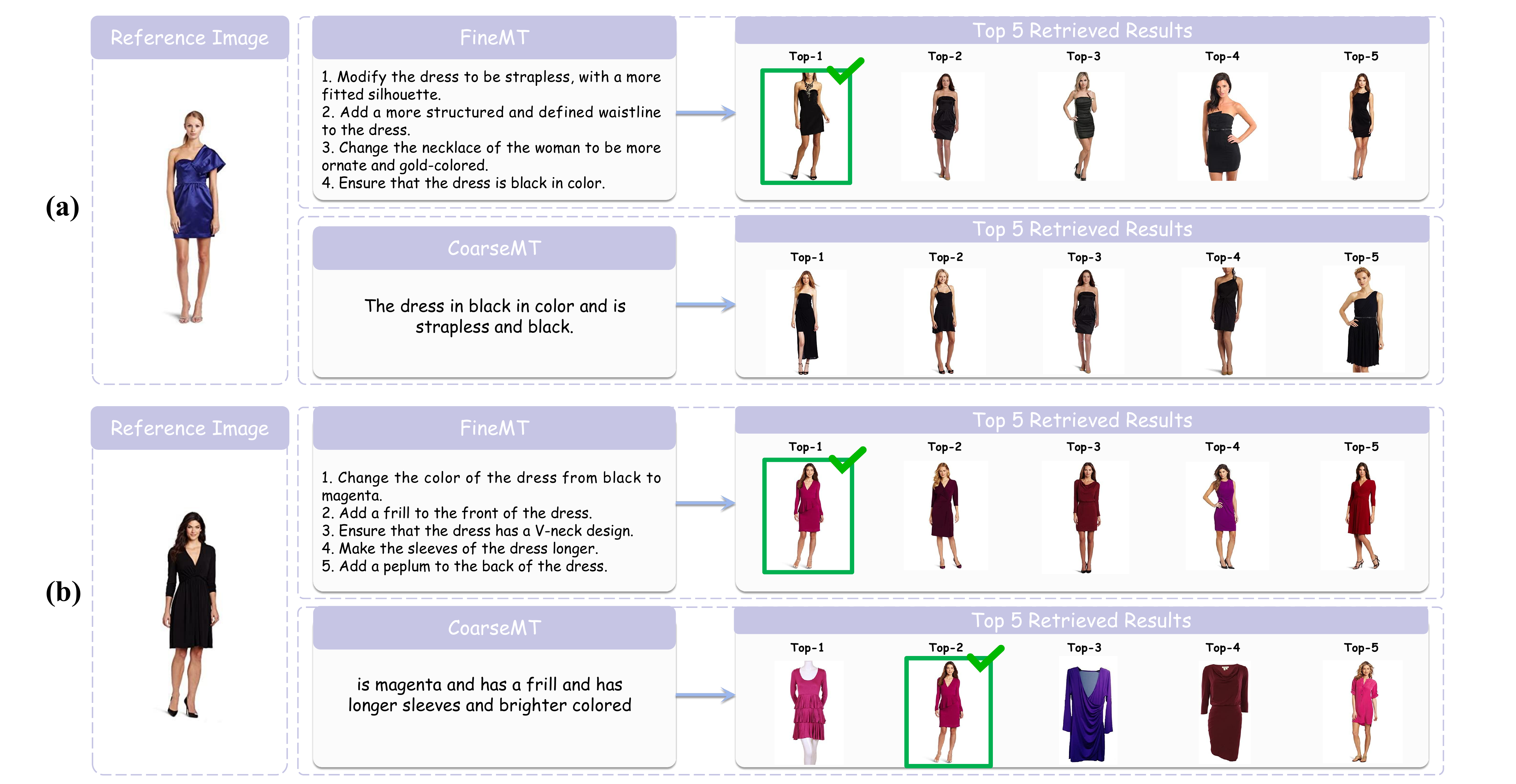}
  \caption{\small The ability to distinguish between visually similar samples using FineMT. We showed the top 5 retrieved results on both Fine-FashionIQ and FashionIQ. The target images are framed in green.}
  \vspace{10pt}
  \label{fig:fiq-sim}
\end{figure*}

\begin{figure*}[ht]
  \centering
  \includegraphics[width=\linewidth]{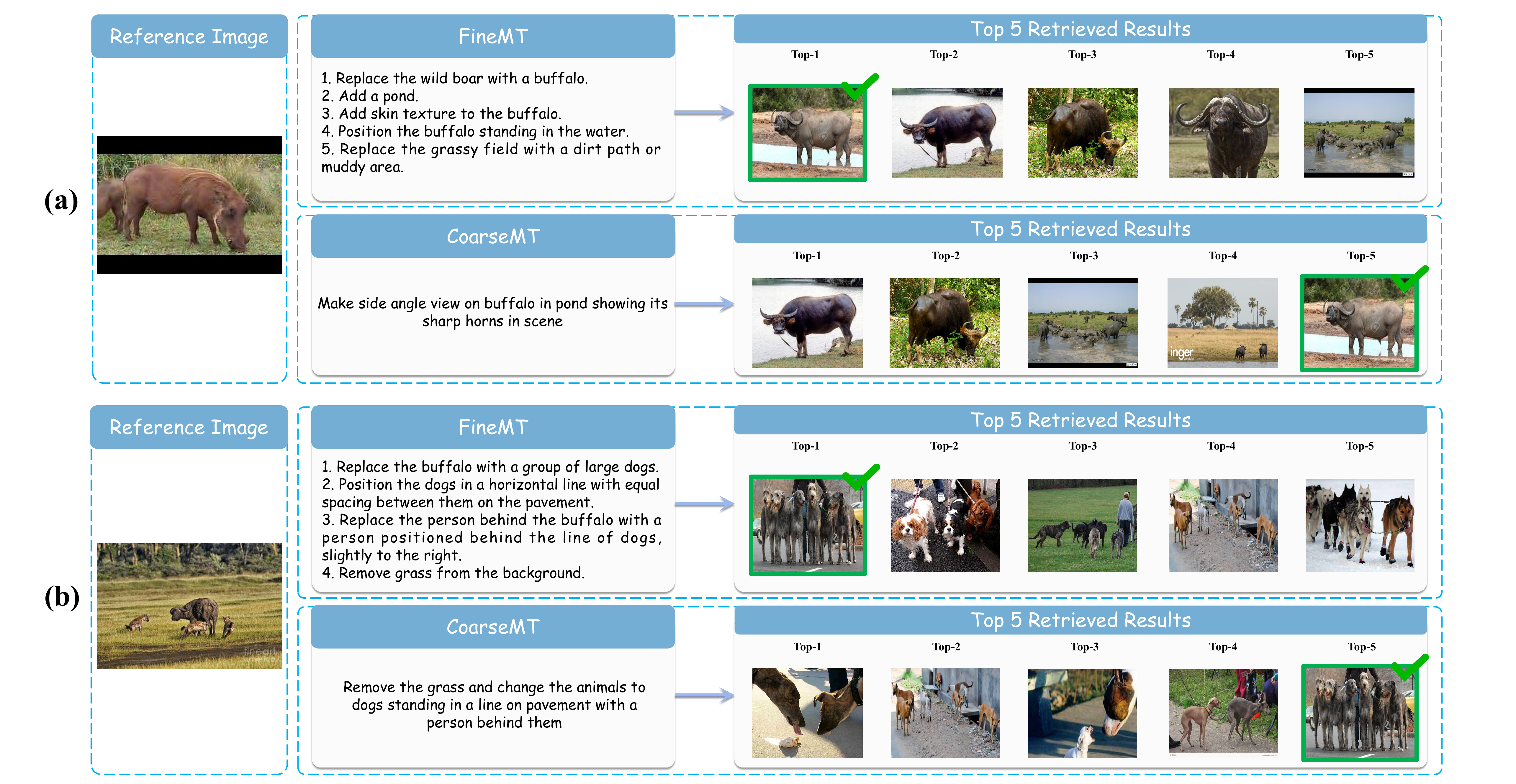}
  \caption{\small The ability to distinguish between visually similar samples using FineMT. We showed the top 5 retrieved results on both Fine-CIRR and CIRR. The target images are framed in green.}
  \vspace{10pt}
  \label{fig:cirr-sim}
\end{figure*}

To validate the advantages of our proposed Fine-FashionIQ and Fine-CIRR datasets in differentiating visually similar samples, we selected BLIP4CIR~\cite{blip4cir}, a straightforward baseline incorporating multimodal query features for retrieving target images. Experiments were conducted comparing FineMT and the original CoarseMT scenarios. Figures~\ref{fig:fiq-sim} and \ref{fig:cirr-sim} illustrate the top-5 retrieval results in both scenarios, highlighting target images in green.

In the fashion-domain dataset Fine-FashionIQ, Figure~\ref{fig:fiq-sim}(a) (bottom) shows that although the top-5 retrieved results satisfy the multimodal query, they are incorrectly classified as negative samples. This occurs because the CoarseMT failed to mention the presence of ``a necklace'' in the target image. After applying FineMT relabeling (top), this detail is captured, correctly identifying the target image as the only positive sample, significantly improving discrimination among visually similar samples. A similar improvement is observed in Figure~\ref{fig:fiq-sim}(b), where the target image's ranking improves from second to first place after FineMT relabeling.

In the open-domain dataset Fine-CIRR, Figure~\ref{fig:cirr-sim}(a) demonstrates that the CoarseMT includes basic descriptions (``side angle view on buffalo'', ``in pond'', and ``sharp horns''), whereas FineMT introduces additional crucial details (``standing in the water'', ``dirt path'', and ``trees in the background''). The enhanced FineMT query accurately retrieves the target image, effectively reducing confusion from visually similar images. Likewise, Figure~\ref{fig:cirr-sim}(b) shows the target image's retrieval rank improving from fifth to first after FineMT relabeling. These results confirm that FineMT's detailed multimodal queries successfully differentiate visually similar samples, converting previously ambiguous cases into true negatives.

\subsection{Comparison of Features}
\label{sup:A.2}
\begin{figure*}[t]
  \centering
  \includegraphics[width=\linewidth]{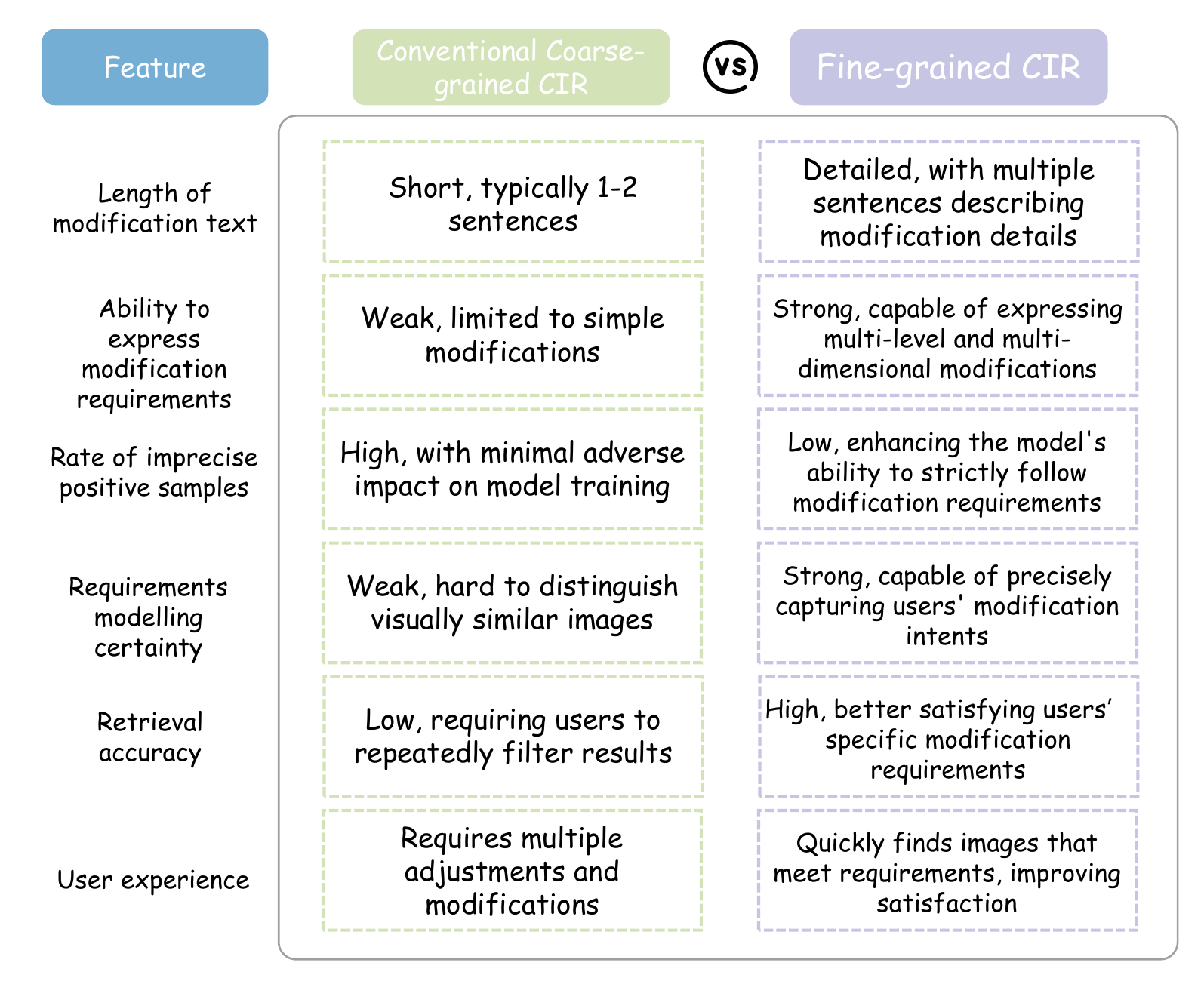}
  \caption{\small Comparison of the characteristics of coarse-grained CIR and fine-grained CIR.}
  \label{fig:sup_cont}
\end{figure*}

As shown in Figure~\ref{fig:sup_cont}, we present the comparison between coarse-grained CIR and fine-grained CIR. 
The table highlights key differences in their ability to handle modification requirements and user intent. 
Coarse-grained CIR typically relies on brief, one-to-two-sentence modification texts, limiting its ability to express complex or nuanced requirements. 
This results in weaker support for multi-dimensional modifications and lower certainty in modeling precise user needs, making it harder to distinguish visually similar images. Additionally, the retrieval accuracy is relatively low, which reduces overall user satisfaction.

In contrast, fine-grained CIR employs detailed modification descriptions, often comprising multiple sentences, to express multi-level and multi-dimensional modifications more effectively. 
This approach reduces the rate of imprecise positive samples, enabling the model to better adhere to specific modification requirements. 
Fine-grained CIR demonstrates stronger certainty in capturing user intent, significantly improving retrieval accuracy by providing results that better align with user expectations. 
Consequently, it enhances the user experience by quickly finding images that meet the desired specifications, reducing the need for iterative adjustments and increasing overall satisfaction. 

\section{Details of Dataset Construction}
\label{sup:B}
In subsection~\ref{sup:B.1},  we provided a more detailed analysis of prompts, including the prompts for LLM in the \textit{Data Selection stage}, the \textit{Dataset Construction stage}, and the \textit{Quality Check stage}. 
In subsection~\ref{sup:B.2}, we analyzed the detailed prompts used for FineMT generation during the process of BLIP-3, which is the key step in the construction process.
And in subsection~\ref{sup:B.3}, we reported the changes in the query number during the construction of our proposed Fine-FashionIQ and Fine-CIRR.

\subsection{Analysis of Prompts for Check}
\label{sup:B.1}
\begin{figure}[t]
  \centering
  \includegraphics[width=\linewidth]{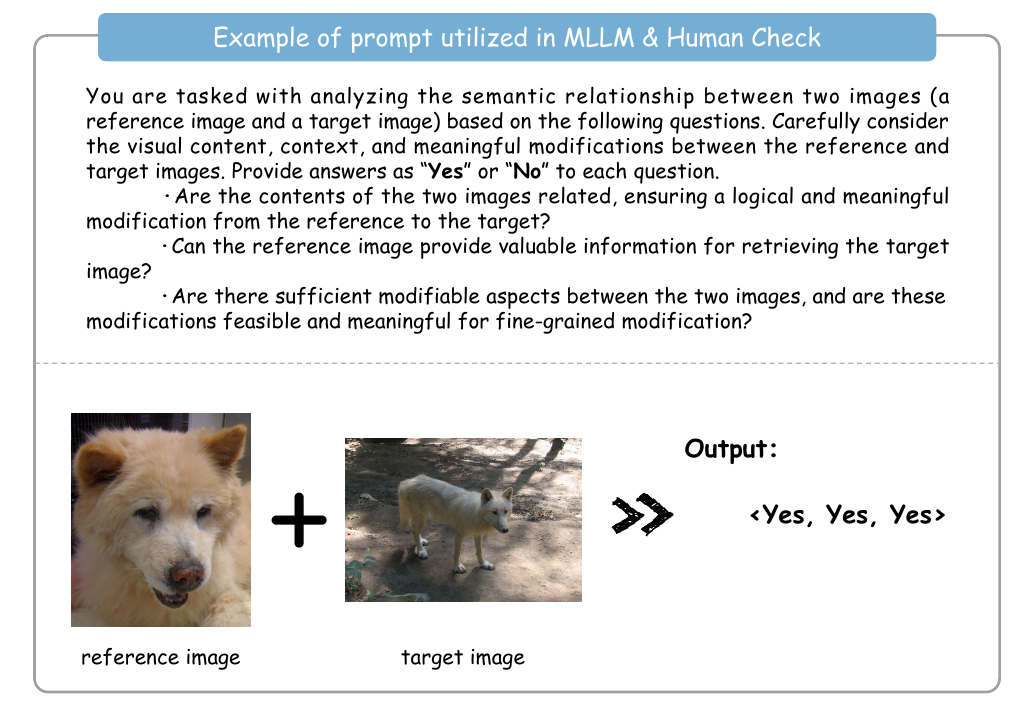}
  \caption{\small Example of prompt utilized in MLLM$\&$Human Check.}
  \label{fig:sup_prompt1}
\end{figure}

Figure~\ref{fig:sup_prompt1} illustrates our designed prompt used in the ``MLLM \& Human Check'' step. 
Here, the LLM analyzes reference-target image pairs based on three provided questions, returning ``Yes'' or ``No'' answers. These answers serve as critical assessments for subsequent evaluation. For instance, as shown, the LLM examines the reference and target images and answers $<$Yes, Yes, Yes$>$. That is because the target can be obtained by modifying details such as the background and subject of the reference image, and the two images are highly related, being dogs of similar breeds, so this pair is well suited for constructing fine-grained datasets.

\begin{figure}[t]
  \centering
  \includegraphics[width=\linewidth]{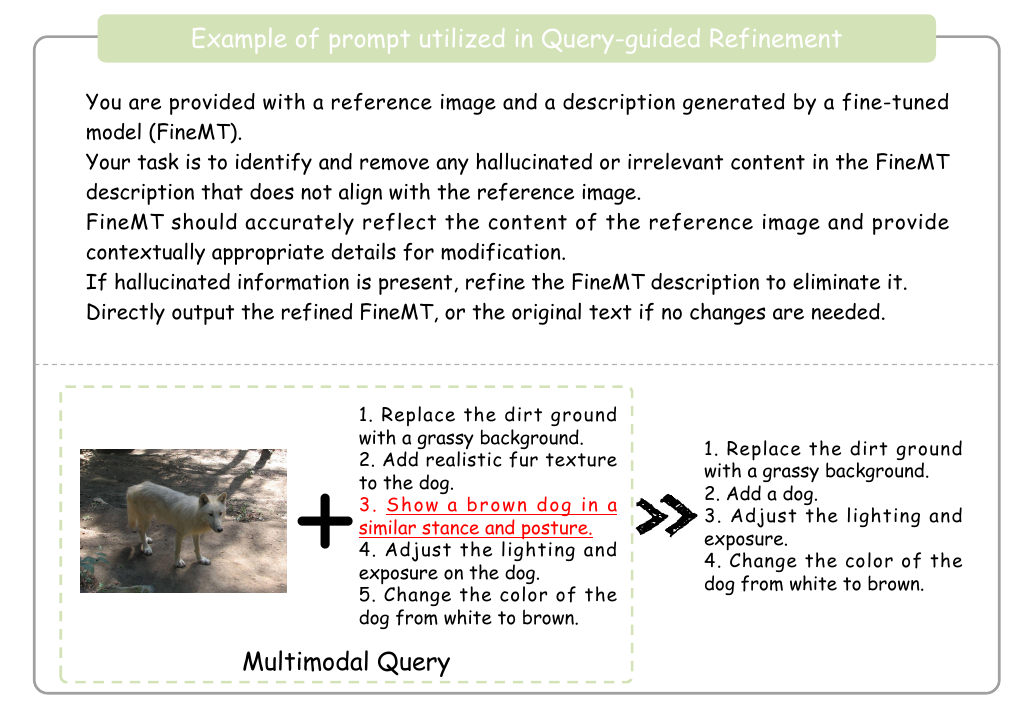}
  \vspace{-10pt}
  \caption{\small Example of prompt utilized in Query-guided Refinement.}
    \vspace{-10pt}
  \label{fig:sup_prompt2}
\end{figure}

Figure~\ref{fig:sup_prompt2} displays the designed prompt for the ``Query-guided Refinement'' step.
Since inputting FineMT alone may omit context regarding the modified objects, we input the entire multimodal query, including the reference image and FineMT, into Llama3.2~\cite{llama-3}. This comprehensive input enables Llama3.2 to accurately interpret modifications in relation to the reference image, thereby ensuring FineMT’s accuracy and eliminating any irrelevant or hallucinated content.
In the provided example, the sentence ``Show a brown dog in a similar stance and posture'' solely describes the target image, making it irrelevant to the reference image and thus a candidate for removal.

\begin{figure}[t]
  \centering
  \includegraphics[width=\linewidth]{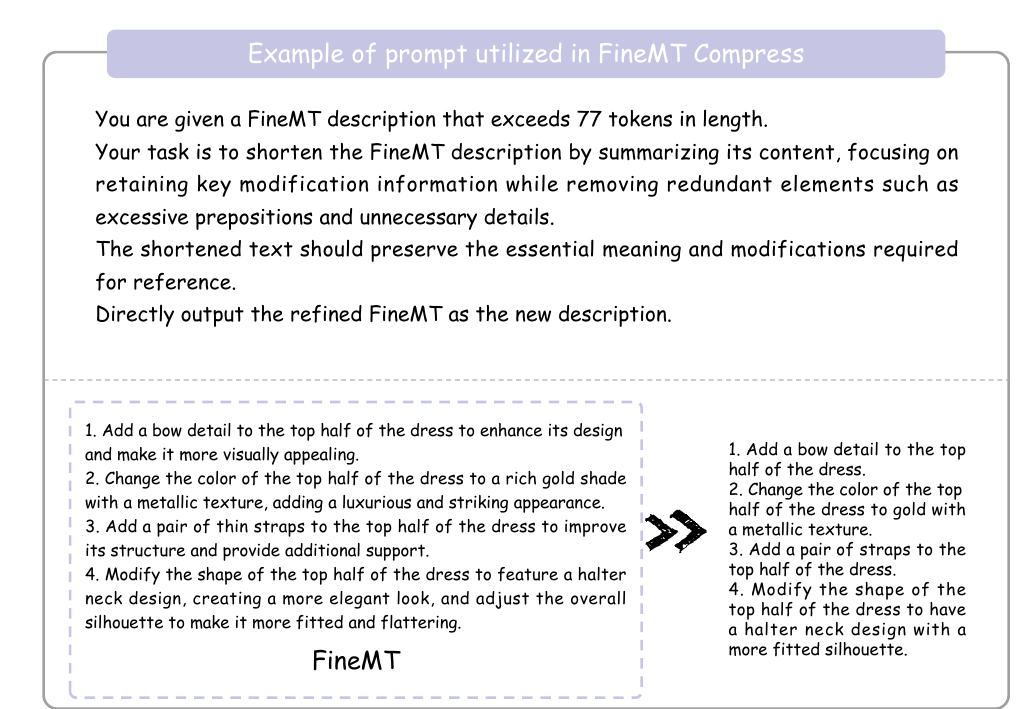}
  \vspace{-10pt}
  \caption{\small Example of prompt utilized in FineMT Compress.}
    \vspace{-10pt}
  \label{fig:sup_prompt3}
\end{figure}

Figure~\ref{fig:sup_prompt3} demonstrates our prompt used in the ``FineMT Compress'' step, which aims to refine textual descriptions by minimizing redundancy while retaining key information. The prompt guides GPT-4o~\cite{gpt4o} in summarizing lengthy FineMT descriptions exceeding 77 tokens. 
This ensures essential modification details are maintained, eliminating excessive prepositions and unnecessary elaboration. The provided example highlights the contrast between verbose input and the refined output, simplifying overly detailed expressions such as ``to enhance its design and make it more visually appealing'' to concise phrases like ``to the top half of the dress''. This compression effectively balances information retention and conciseness, thus enhancing the quality of FineMT.

\begin{figure*}[t]
  \centering
  \includegraphics[width=\linewidth]{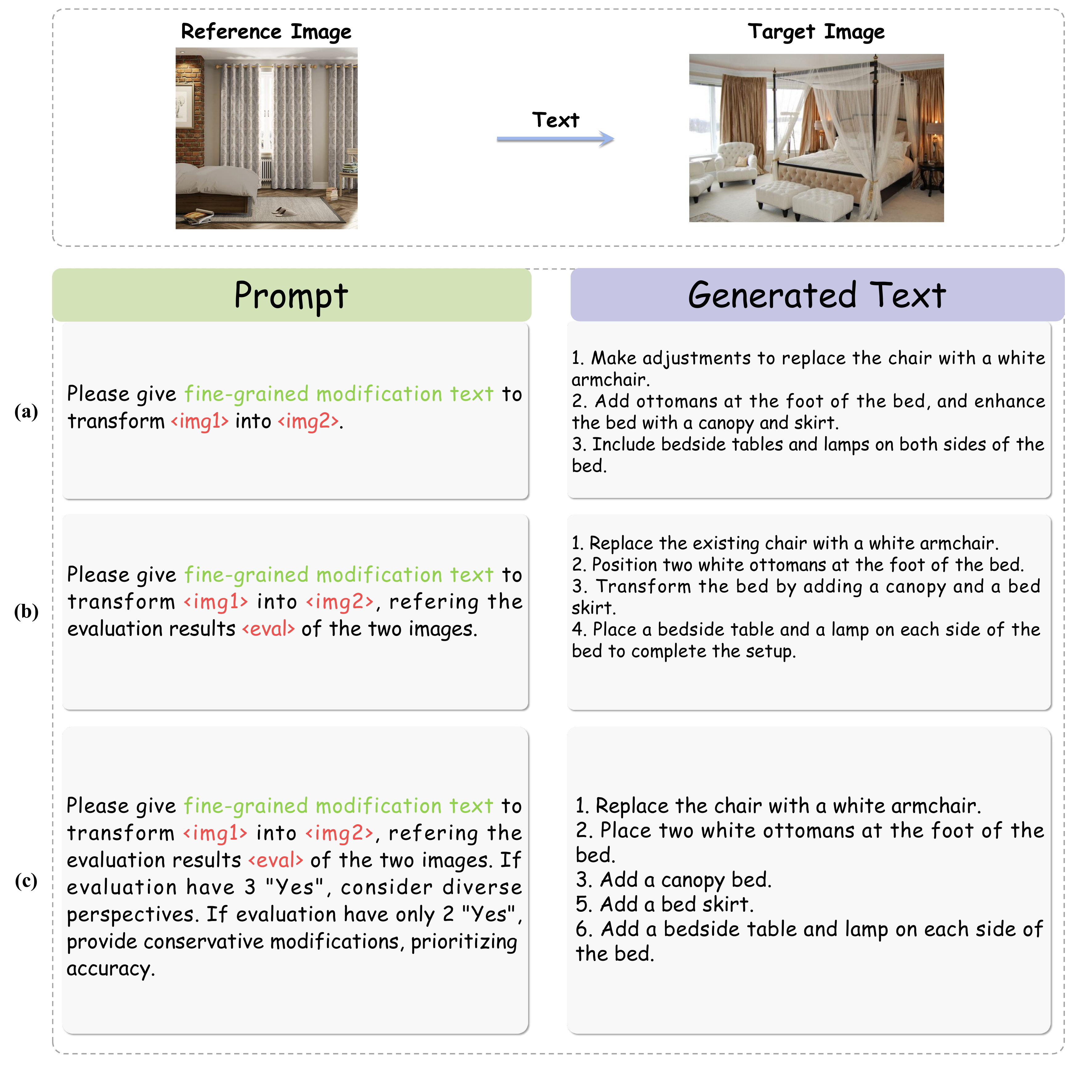}
  \vspace{-35pt}
  \caption{\small Generated texts using various prompts for BLIP-3.}
  \vspace{10pt}
  \label{fig:blip3-prompt}
\end{figure*}

\begin{figure*}[t]
  \centering
  \includegraphics[width=\linewidth]{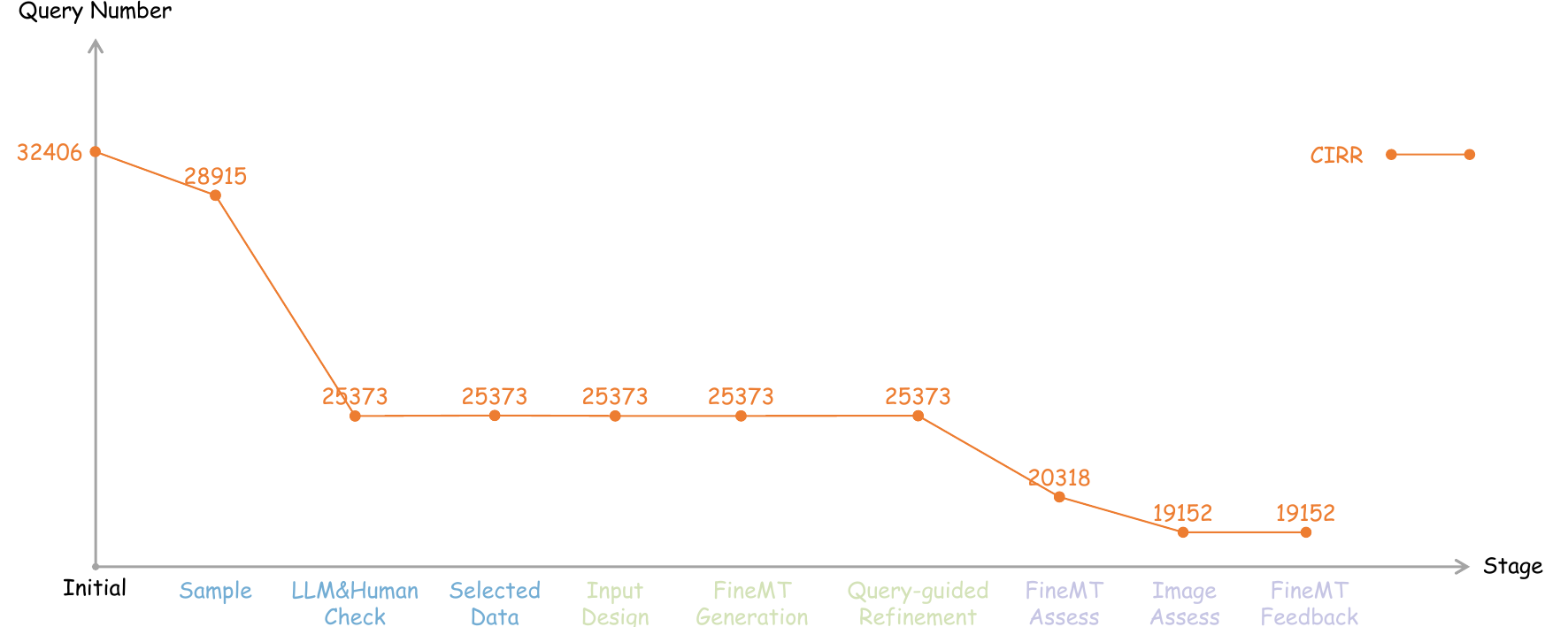}
  \caption{\small Changes in the query number throughout the construction of Fine-CIRR.}
  \label{fig:cirr-num}
\end{figure*}

\begin{figure*}[t]
  \centering
  \includegraphics[width=\linewidth]{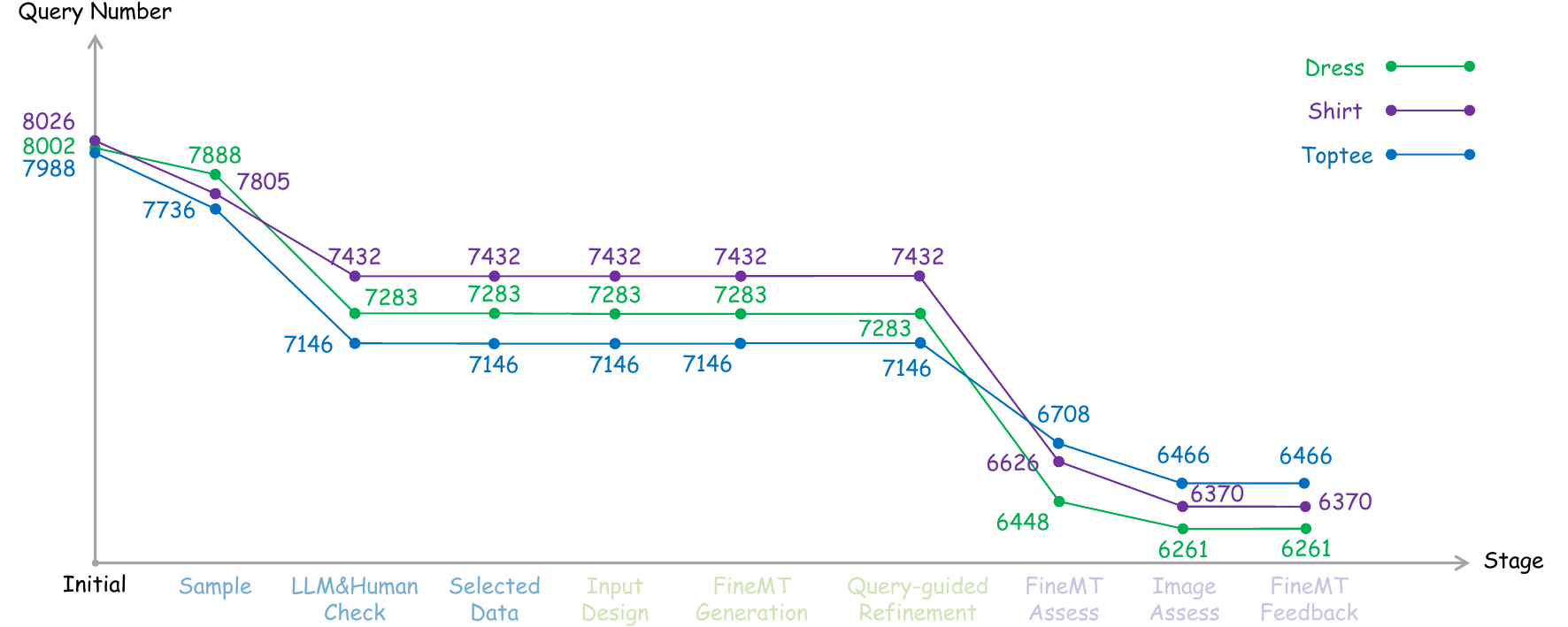}
  \caption{\small Changes in the query number throughout the construction of Fine-FashionIQ.}
  \label{fig:fiq-num}
\end{figure*}

\subsection{Analysis of Prompts for FineMT Generation}
\label{sup:B.2}
In Figure~\ref{fig:blip3-prompt}, we present the prompts employed to generate foundational texts for FineMT using BLIP-3~\cite{blip3}. Our requirement was that BLIP-3 outputs include additional details present in the image pairs but omitted in the original CoarseMT. For instance, while the original CoarseMT might only reference a single object, the target image typically reflects modifications across multiple objects relative to the reference image. Thus, FineMT provides a comprehensive and detailed description of the modifications, addressing the shortcomings of the original CoarseMT.

or a thorough analysis, we compare generated texts obtained using various designed prompts for BLIP-3, as illustrated in Figure~\ref{fig:blip3-prompt}. In this comparison, prompt (a), which lacks evaluation by an LLM, generates texts limited in detail, failing to capture comprehensive modifications. Conversely, prompts (b) and (c), aided by LLM evaluation, successfully capture and elaborate on fine-grained modifications from diverse perspectives. This demonstrates the effectiveness of incorporating LLM evaluation to enhance the accuracy and completeness of captured fine-grained details.

Furthermore, texts generated using prompt (c) were notably more precise and comprehensive, addressing various perspectives and details effectively. Therefore, we selected prompt (c) as the optimal approach for generating fine-grained texts.

Recognizing differences between the fashion-domain data and real-life scenarios, we employed specifically designed additional prompts for Fine-FashionIQ. Given the original FashionIQ dataset's focus on garment details, such as the presence or absence of straps and sleeve lengths—these tailored prompts guide BLIP-3's output towards meticulous attention to garment components, holistic interpretation of clothing items, and comparative analysis between reference and target images. This meticulous approach enables the generation of highly detailed modification descriptions.

In contrast, Fine-CIRR presents a more complex challenge due to the open-domain nature of the original CIRR dataset, typically involving multiple objects and background elements. This complexity underscores the essential role of FineMT, whose detailed descriptions precisely capture the genuine modifications present in the reference and target images, further validating the need for enhanced granularity.

\subsection{Query Number Changes throughout the Construction}
\label{sup:B.3}
Furthermore, since the fine-grained datasets we constructed have a large number of checking and filtering steps, we statistically recorded the changes in the query number throughout the dataset construction process and presented it as line graphs, as shown in Figure~\ref{fig:cirr-num} and Figure~\ref{fig:fiq-num}.

\begin{table*}[htbp]
  \centering

    \begin{tabular}{lllllcccccccc}
    \hline
    \multicolumn{2}{c}{\multirow{2}{*}{}} & \multicolumn{1}{c}{\multirow{2}{*}{Domain}} & \multicolumn{2}{c}{\multirow{2}{*}{Granularity}} & \multicolumn{2}{c}{\multirow{2}{*}{Train triplets}} & \multicolumn{2}{c}{\multirow{2}{*}{Test triplets}} & \multicolumn{2}{c}{\multirow{2}{*}{MT length$^{\textcolor{red}{*}}$}} & \multicolumn{2}{c}{\multirow{2}{*}{Maximal tokens}} \\
    \multicolumn{2}{c}{} &       & \multicolumn{2}{c}{} & \multicolumn{2}{c}{} & \multicolumn{2}{c}{} & \multicolumn{2}{c}{} & \multicolumn{2}{c}{} \\
    \hline
    \multicolumn{2}{l}{FashionIQ} & fashion & \multicolumn{2}{l}{coarse-grained} & \multicolumn{2}{c}{18000} & \multicolumn{2}{c}{6016} & \multicolumn{2}{c}{24.7} & \multicolumn{2}{c}{37.0} \\
    \multicolumn{2}{l}{FineFashionIQ} & fashion & \multicolumn{2}{l}{fine-grained} & \multicolumn{2}{c}{14812} & \multicolumn{2}{c}{4285} & \multicolumn{2}{c}{59.2} & \multicolumn{2}{c}{77.0} \\
    \hline
    \multicolumn{2}{l}{CIRR} & open  & \multicolumn{2}{l}{coarse-grained} & \multicolumn{2}{c}{28225} & \multicolumn{2}{c}{4181} & \multicolumn{2}{c}{12.8} & \multicolumn{2}{c}{50.0} \\
    \multicolumn{2}{l}{FineCIRR} & open  & \multicolumn{2}{l}{fine-grained} & \multicolumn{2}{c}{15344} & \multicolumn{2}{c}{3808} & \multicolumn{2}{c}{62.9} & \multicolumn{2}{c}{77.0} \\
    \hline
    \end{tabular}%
  \caption{Statistics of fine-grained datasets for CIR compared to two classic CIR datasets, FashionIQ and CIRR. MT length$^{\textcolor{red}{*}}$ denotes the average length of modification texts in the dataset.}
  \label{tab:lens}%
\end{table*}%

\section{Traditional and Fine-grained CIR Datasets}
\label{sup:C}
In subsection~\ref{sup:C.1}, we present our proposed datasets and their underlying datasets, FashionIQ~\cite{FashionIQ} and CIRR~\cite{cirr}.
In subsection~\ref{sup:C.2}, we compared and analyzed the statistics of the datasets.
\subsection{Dataset Details}
\label{sup:C.1}
To evaluate the validity of the fine-grained modification text, we constructed two fine-grained datasets for CIR. We now describe each dataset in detail as follows,
\begin{itemize}
     \item \textbf{Fine-FashionIQ} is based on the classic CIR dataset, \textbf{FashionIQ}~\cite{FashionIQ}, whose content belongs entirely to the fashion domain. It consists of $77,684$ images which are divided into three categories: \textit{Dresses}, \textit{Shirts}, and \textit{Tops\&Tees}. Following the FashionIQ, we treat it as three independent datasets. And there are $18$K triples for training and \mbox{$\sim$\hspace{0em}$6$K} triples for testing.
     Based on it, after our detailed filtering and constructing process, we finally utilized $14812$ and $4285$ triplets for training and testing on Fine-FashionIQ.
    \item \textbf{Fine-CIRR} is based on the classic open-domain CIR dataset, \textbf{CIRR}~\cite{cirr}. It contains $21,552$ real images taken from the renowned language reasoning dataset ${\operatorname{NLVR}}^2$, which is well-known for its natural language reasoning applications. CIRR contains a total of $36.5$ K triples, of which $80$\% are used for training, $10$\% for validation and $10$\% for testing.
    Specifically, similar to Fine-FashionIQ, we utilizing a detailed process and finally obtained $15344$ and $3808$ triplets for training and testing on Fine-CIRR. In addition, following the CIRR, the fine-grained dataset Fine-CIRR includes a specialized subset designed for fine discrimination. This subset focuses on negative images that exhibit a high degree of visual similarity and is utilized to assess the model's performance in distinguishing false-negative images.
\end{itemize}

\begin{table*}[ht]
  \centering
           \resizebox{\linewidth}{!}{
    \begin{tabular}{l|l|cc|cc|cc|cc|cccc|ccc|c}
\Xhline{2pt}
    \multicolumn{1}{c|}{\multirow{3}{*}{Method}} & \multicolumn{1}{c|}{\multirow{3}{*}{Year}} & \multicolumn{8}{c|}{FashionIQ}                                & \multicolumn{8}{c}{CIRR} \\
\cline{3-18}          &       & \multicolumn{2}{c|}{Dresses} & \multicolumn{2}{c|}{Shirts} & \multicolumn{2}{c|}{Tops\&Tees} & \multicolumn{2}{c|}{Avg} & \multicolumn{4}{c|}{R@k}      & \multicolumn{3}{c|}{Rsubset@k} & \multirow{2}{*}{Avg} \\
\cline{3-17}          &       & R@10  & R@50  & R@10  & R@50  & R@10  & R@50  & R@10  & R@50  & k=1   & k=5   & k=10  & k=50  & k=1   & k=2   & k=3   &  \\
    \hline
    \hline
    TIRG~\cite{tirg}  & \multicolumn{1}{c|}{2019} & 14.87  & 34.66  & 18.26  & 37.89  & 19.08  & 39.62  & 17.40  & 37.39  & 14.61 & 48.37 & 64.08 & 90.03 & 22.67 & 44.97 & 65.14 & 35.52 \\
    CLVC-Net~\cite{clvcnet} & \multicolumn{1}{c|}{2021} & 29.85  & 56.47  & 28.75  & 54.76  & 33.50  & 64.00  & 30.70  & 58.41  & -     & -     & -     & -     & -     & -     & -     & - \\
    CLIP4CIR~\cite{clip4cir-v2} & \multicolumn{1}{c|}{2022} & 33.81  & 59.40  & 39.99  & 60.45  & 41.41  & 65.37  & 38.40  & 61.74  & 38.53 & 69.98 & 81.86 & 95.93 & 68.19 & 85.64 & 94.17 & 69.09 \\
    TG-CIR~\cite{tgcir} & \multicolumn{1}{c|}{2023} & 45.22  & 69.66  & 52.60  & 72.52  & 56.14  & 77.10  & 51.32  & 73.09  & 45.25 & 78.29 & 87.16 & 97.30 & 72.84 & 89.25 & 95.13 & 75.57 \\
    BLIP4CIR~\cite{blip4cir} & \multicolumn{1}{c|}{2024} & 40.65  & 66.34  & 40.38  & 64.13  & 46.86  & 69.91  & 42.63  & 66.79  & 40.17 & 71.81 & 83.18 & 95.69 & 72.34 & 88.70 & 95.23 & 72.08 \\
    CoVR-2~\cite{covr-2} & \multicolumn{1}{c|}{2024} & 46.53  & 69.60  & 51.23  & 70.64  & 52.14  & 73.27  & 49.96  & 71.17  & 50.43 & 81.08 & 88.89 & \underline{98.05} & 76.75 & 90.34 & 95.78 & 78.92 \\
    Candidate~\cite{candidate} & \multicolumn{1}{c|}{2024} & 48.14  & 71.34  & 50.15  & 71.25  & 55.23  & 76.80  & 51.17  & 73.13  & 50.55 & 81.75 & \underline{89.78} & 97.18 & 80.04 & 91.90 & 96.58 & 80.90 \\
    SPRC~\cite{sprc}  & \multicolumn{1}{c|}{2024} & 49.18  & 72.43  & 55.64  & 73.89  & 59.35  & 78.58  & 54.92  & 74.97  & \underline{51.96}  & \underline{82.12}  & 89.74  & 97.69  & \textbf{80.65}  & \underline{92.31}  & \underline{96.60}  & \underline{81.39} \\
    LIMN~\cite{limn}  & \multicolumn{1}{c|}{2024} & \underline{50.72}  & \underline{74.52}  & \underline{56.08}  & \underline{77.09}  & \underline{60.94}  & \underline{81.85}  & \underline{55.91}  & \underline{77.82}  & 43.64 & 75.37 & 85.42 & 97.04 & 69.01 & 86.22 & 94.19 & 72.19 \\
    IUDC~\cite{iudc} & \multicolumn{1}{c|}{2025} & 35.22  & 61.90  & 41.86  & 63.52  & 42.19  & 69.23  & 39.76  & 64.88  & -     & -     & -     & -     & -     & -     & -     & -  \\
    \hline
    \hline
                    \rowcolor[rgb]{ .851,  .851,  .851}
    \textbf{FineCIR (Ours)} &   &  \textbf{53.15} & 	\textbf{76.95} 	& \textbf{63.98} & 	\textbf{81.35} &	\textbf{62.45} & 	\textbf{82.98} & 	\textbf{59.86} 	& \textbf{80.43}   & \textbf{52.75 } & \textbf{82.96 } & \textbf{90.84 } & \textbf{98.27 } & \underline{80.29}  & \textbf{92.39 } & \textbf{96.70 } & \textbf{81.63 } \\
\Xhline{2pt}
    \end{tabular}%
    }
            \vspace{-10pt}
      \caption{Performance comparison on original FashionIQ and CIRR relative to R@K(\%). The overall best results are in bold, while the best results over baselines are underlined. The Avg metric in CIRR denotes (R@$5$ + R$_{subset}$@$1$) / 2.}
        \vspace{-10pt}
  \label{tab:full_cir}%
\end{table*}%

\subsection{Statistics}
\label{sup:C.2}
We provide detailed statistics for Fine-FashionIQ and Fine-CIRR, comparing them with their original counterparts, FashionIQ and CIRR. 
Since previous annotations led to portions of the data being unsuitable for constructing fine-grained datasets, we performed iterative checks and filtration, resulting in a streamlined dataset with fewer but higher-quality queries.
Regarding the length of modification texts, FineMT inherently contains finer-grained descriptions, thus resulting in increased length compared to the original modification texts, as demonstrated in Table~\ref{tab:lens}. 
Nevertheless, we controlled the length during dataset construction to avoid text overpowering the visual information and weakening image relevance. 
Additionally, controlling text length ensures compatibility with the CLIP~\cite{clip} encoder, which has a maximum input token length limitation of 77 tokens.

\section{Detailed Comparison on Traditional CIR Datasets}
\label{sup:D}
As illustrated in~\Cref{tab:full_cir}, we conduct a comprehensive evaluation of the proposed FineCIR model against several significant baselines on the traditional CIR datasets, \ie, FashionIQ and CIRR. From the results presented in the table, we can obtain the following observations.
\textbf{1)} Our proposed FineCIR achieves outstanding performance on both FashionIQ and CIRR, demonstrating its ability to capture essential modification semantics even in coarse-grained modification text.
\textbf{2)} Compared to the baselines, FineCIR exhibits significant performance advantages on both fashion-domain and open-domain datasets. This may be attributed to the superior domain generalization ability of our designed modules.
\textbf{3)} Notably, FineCIR achieves better results on fine-grained CIR datasets compared to traditional CIR datasets, whereas previous baselines (\eg, LIMN, BLIP4CIR) exhibit significant performance drops on fine-grained CIR. This suggests that prior methods struggle to capture fine-grained modification semantics, leading to the lack of critical modification details and performance degradation. In contrast, FineCIR demonstrates superior generalization capability in capturing modification semantics across both coarse and fine granularities.

\section{Computation Cost Analysis}
\label{sup:E}
We determine the computational costs of our proposed FineCIR compared to the representative CIR models, Candidate~\cite{candidate} and SPRC~\cite{sprc}, on the Fine-CIRR dataset. Specifically, we choose the train time, test time, and GPU memory for evaluation, as shown in~\Cref{tab:cost}. 
All experiments are performed on a single NVIDIA A40 GPU. 
The train time describes the time it takes for the model to optimize to the optimum, while the test time is the time it takes for the inference on one sample. 
From the results presented in~\Cref{tab:cost}, we observe that FineCIR's \textbf{w/o SG} setting obtains optimal training and testing efficiency, and it has a higher recall than Candidate and SPRC. Importantly, this study prioritizes enhanced retrieval accuracy through improved model architecture and more effective input information handling, rather than targeting scenarios where speed is prioritized over precision. Consequently, we accept a modest increase in computational overhead to gain up to an $11.4$\% absolute recall improvement compared to SPRC on fine-grained CIR benchmarks.

\begin{table}[ht]
  \centering
    \vspace{-6pt}

  \vspace{-3pt}
            \resizebox{\linewidth}{!}{
    \begin{tabular}{cc|cc|cc|cc|c|c}
    \Xhline{2pt}
    \multicolumn{2}{c|}{Method} & \multicolumn{2}{c|}{Backbone} & \multicolumn{2}{c|}{Train$\,\,$$\downarrow$} & \multicolumn{2}{c|}{Test$\,\,$$\downarrow$} & \multicolumn{1}{c}{Memory$\,\,$$\downarrow$} & \multicolumn{1}{c|}{Avg.R$\,\,$$\uparrow$}\\
    \hline
    \multicolumn{2}{l|}{Candidate} & \multicolumn{2}{c|}{BLIP} & \multicolumn{2}{c|}{16h} & \multicolumn{2}{c|}{\underline{16.7ms/sample}} &47.3G & 78.09 \\
    
    \multicolumn{2}{l|}{SPRC} & \multicolumn{2}{c|}{BLIP-2} &\multicolumn{2}{c|}{4.5h} & \multicolumn{2}{c|}{17.7ms/sample}  &\textbf{25.2G} & 80.97\\
    \rowcolor[rgb]{ .851,  .851,  .851}
    \multicolumn{2}{l|}{\textbf{FineCIR  w/o SG}} & \multicolumn{2}{c|}{BLIP-2}  & \multicolumn{2}{c|}{\textbf{2.7h}} &  \multicolumn{2}{c|}{\textbf{15.6 ms/sample}} &\underline{25.4G} & \underline{81.12}\\
                \rowcolor[rgb]{ .851,  .851,  .851}
    \multicolumn{2}{l|}{\textbf{FineCIR (Ours)}} & \multicolumn{2}{c|}{BLIP-2}   & \multicolumn{2}{c|}{\underline{2.8h}} &  \multicolumn{2}{c|}{23.1 ms/sample} &28.9G & \textbf{84.73}\\
    \Xhline{2pt}

    \end{tabular}%
    }
  \caption{Comparison on computation cost. The better results are in bold.}
  \label{tab:cost}%
\end{table}%

\section{Case Study}
\label{sup:F}
\begin{figure*}[t]
  \centering
  \includegraphics[width=\linewidth]{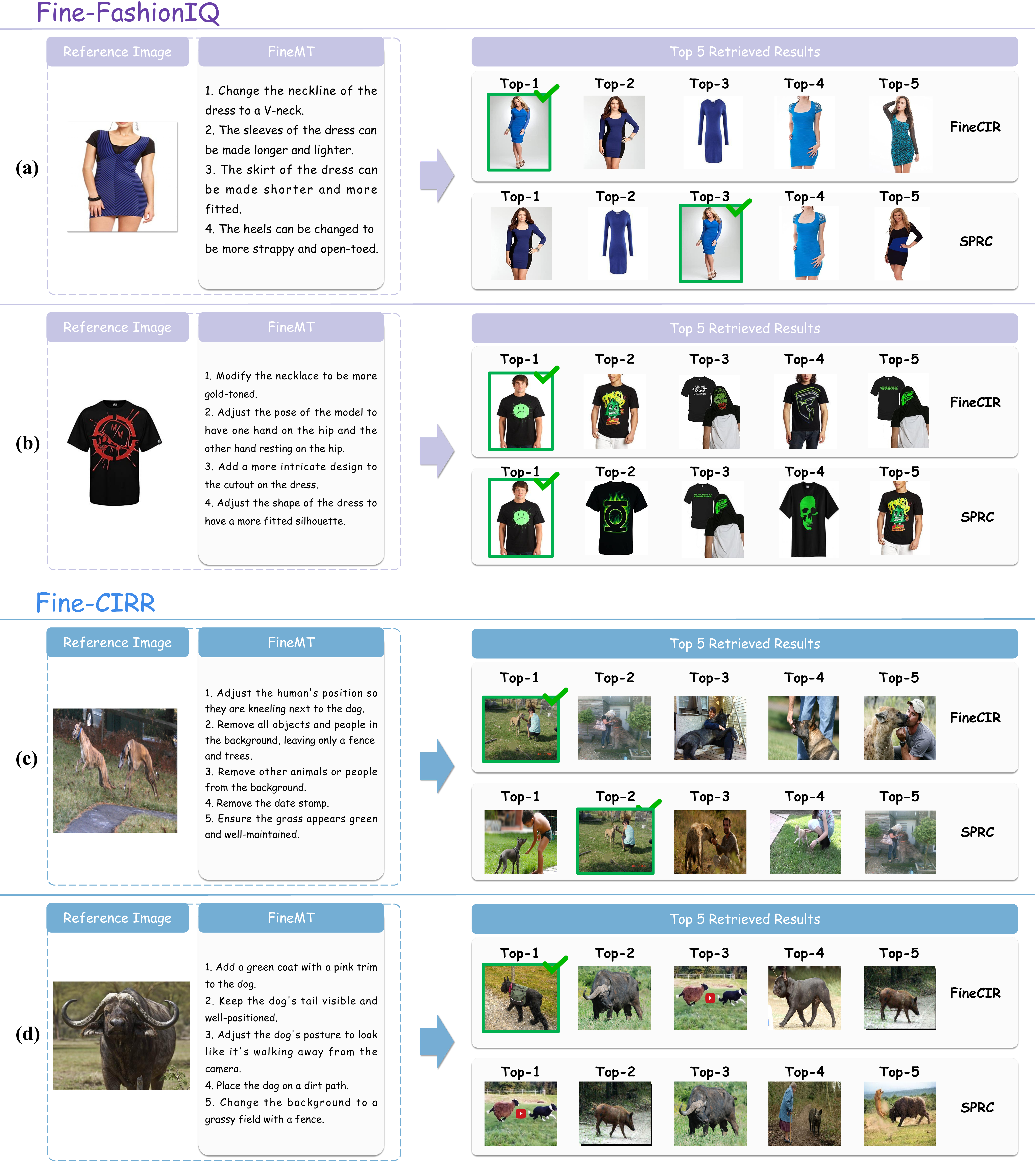}
  \vspace{-20pt}
  \caption{\small Qualitative examples of our proposed FineCIR compared to the sub-optimal CIR model SPRC.}
    \vspace{-10pt}
  \label{fig:case}
\end{figure*}

To intuitively validate the performance of our proposed FineCIR on fine-grained scenarios of CIR, we present several examples demonstrating FineCIR's retrieval results to this scenario, along with the comparison to sub-optimal CIR model SPRC~\cite{sprc}, as shown in Figure~\ref{fig:case}.
In the Fine-FashionIQ dataset, Figure~\ref{fig:case}(a) shows that FineCIR accurately retrieves the target image at Top-1 by capturing fine-grained modifications to the dress, including changes to the neckline, sleeves, skirt, and heels, while SPRC ranks the target at Top-3, reflecting a less precise understanding of the modifications. 
Similarly, in Figure~\ref{fig:case}(b), FineCIR retrieves the target image at Top-1 for modifications involving the necklace, pose, and design details of a shirt, whereas SPRC ranks it at Top-1 too, demonstrating both models' ability to prioritize subtle changes. 
On the Fine-CIRR dataset, Figure~\ref{fig:case}(c) illustrates FineCIR's success in retrieving the target image at Top-1 based on modifications like repositioning a person and refining background elements, while SPRC ranks the target at Top-2. 
In Figure~\ref{fig:case}(d), FineCIR also ranks the target at Top-1 by accurately addressing modifications to the dog's position, posture, and background, in contrast to SPRC's out of Top 5 ranking. 
These results demonstrate that FineCIR consistently outperforms SPRC by effectively capturing and prioritizing fine-grained semantics of modifications across both fashion and open-domain CIR tasks.
\end{document}